\documentclass[runningheads]{llncs}

\usepackage[T1]{fontenc}

\usepackage{eccvabbrv}

\usepackage{graphicx}
\usepackage{booktabs}
\usepackage{comment}
\usepackage{amsmath,amssymb}
\usepackage{color}
\usepackage{url}

\usepackage{subcaption}
\usepackage{rotating}
\usepackage{multirow}
\usepackage{tabularray}
\usepackage{array}
\usepackage[
  locale = US,
]{siunitx}
\DeclareSIUnit{\pp}{\textup{pp}}
\usepackage[acronym]{glossaries}
\setacronymstyle{long-short}

\usepackage{calc}
\usepackage{colortbl}
\usepackage{pgfplotstable}
\usetikzlibrary{pgfplots.colorbrewer}
\pgfplotsset{compat=1.18}
\newcommand{\ra}[1]{\renewcommand{\arraystretch}{#1}}
\newcommand{\textover}[3][l]{%
 \makebox[\widthof{#3}][#1]{#2}%
}

\newcommand{\rotside}[1]{\rotatebox[origin=c]{90}{#1}}

\usepackage{hyperref}

\usepackage[capitalize]{cleveref}
\crefname{section}{Sec.}{Secs.}
\Crefname{section}{Section}{Sections}
\crefname{table}{Tab.}{Tabs.}
\Crefname{table}{Table}{Tables}

\newif\ifreview
\reviewfalse

\ifreview
	\usepackage{lineno}

	\linenumbers
\fi

\newacronym{bn}{BN}{batch normalization}

\begin{document}


\def\SubNumber{17}

\def\GCPRTrack{Main Track}

\title{BTSeg: Barlow Twins Regularization for Domain Adaptation in Semantic Segmentation}

\ifreview
\else
	\titlerunning{BTSeg for Domain Adaptation in Semantic Segmentation}

	\author{Johannes Künzel\inst{1}\orcidID{0000-0002-3561-2758} \and
	Anna Hilsmann\inst{1}\orcidID{0000-0002-2086-0951} \and
	Peter Eisert\inst{1,2}\orcidID{0000-0001-8378-4805}}
	
	\authorrunning{J. Künzel et al.}
	
	\institute{Fraunhofer Heinrich-Hertz-Institut, HHI, Germany \and
        Humboldt University Berlin, Germany
}

\fi

\maketitle              

\begin{abstract}
    We introduce BTSeg (Barlow Twins regularized Segmentation), an innovative, semi-supervised training approach enhancing semantic segmentation models in order to effectively tackle adverse weather conditions without requiring additional labeled training data.
    Images captured at similar locations but under varying adverse conditions are regarded as manifold representation of the same scene, thereby enabling the model to conceptualize its understanding of the environment.    
    BTSeg shows cutting-edge performance for the new challenging ACG benchmark and sets a new state-of-the-art for weakly-supervised domain adaptation for the ACDC dataset.
    To support further research, we have made our code publicly available at \url{https://github.com/fraunhoferhhi/BTSeg}.
\end{abstract}
\section{Introduction}

In autonomous driving, semantic segmentation encounters heightened challenges in adverse conditions, including snow, rain, fog, and darkness.
Precisely annotating images pixel-by-pixel under such conditions is costly and difficult; consider, for example, the challenge of distinguishing between a road and a sidewalk in snowy conditions.
Recent datasets \cite{Sakaridis2021,sakaridis2019} contain pairs of images, taken under adverse and clear weather respectively, roughly aligned using their respective GPS coordinates.

\begin{figure}[ht]
    \centering
    \includegraphics[width=0.9\linewidth]{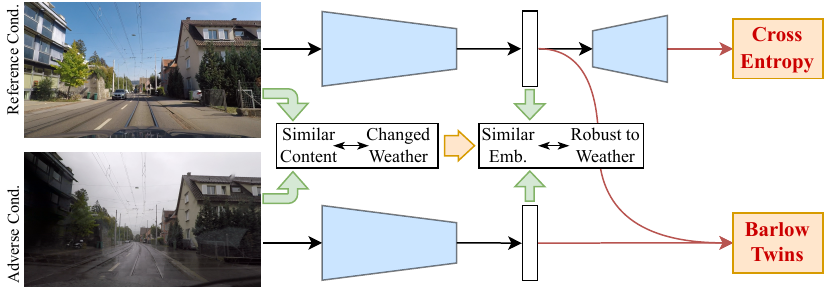} 

    \caption{BTSeg processes two images captured at (approximately) the same location under different weather conditions. This enables the network to learn encodings that not only accurately estimate semantic segmentation but also withstand appearance variations due to changing weather.}
    \label{fig:teaser}
\end{figure}

Taking advantage of these datasets, we present BTSeg (Barlow Twins regularized Segmentation), a novel semi-supervised training strategy for learning semantic segmentation models, which are robust to varying weather conditions, visualized in \cref{fig:teaser}.
BTSeg treats the roughly matched image pairs depicting different weather conditions as natural augmentations of the same underlying scene, thereby forcing the encoder to learn features robust to changing weather conditions.
In consequence, the image pairs are leveraged to teach the network's encoder to find and use the latent information shared by both scenarios.

We argue that generating high-quality labeled ground truth data is easier for clear weather images. 
Thus, BTSeg incorporates their annotations into a second objective function to compel the encoder to produce feature maps suitable for generating high-quality segmentations. 

In summary, our main contributions are the following:

\begin{enumerate}
    \item We introduce BTSeg, a semi-supervised learning strategy, for increased robustness under adverse conditions. This eliminates the need for elaborately annotated bad weather conditions or negative samples as in earlier contrastive methods, such as CMA~\cite{Bruggemann2023CMA}.
    \item We pioneer the integration of Barlow Twins~\cite{zbontar2021} with natural and only coarsely paired images, by offering effective solutions for handling moving objects and small misalignments.   
    \item We demonstrate how our approach can easily be integrated into a state-of-the-art semantic segmentation network, consisting of an encoder and a decoder part, while introducing only a negligibly increase in the number of parameters.
    \item Our approach shows enhanced results when compared against current state-of-the-art methods, setting a new state-of-the-art for the challenging ACG-benchmark, indicating both its efficiency and effectiveness.
\end{enumerate}

\section{Related Work}
\label{sec:relatedwork}

In this section, we briefly review the state-of-the-art in image segmentation under adverse conditions, self-supervised learning and combination thereof. 

\subsection{Weakly-supervised Domain Adaption}

With the introduction of datasets like Dark~Zurich~\cite{sakaridis2019} and ACDC~\cite{Sakaridis2021}, the field of semantic image segmentation has expanded to include adverse weather conditions.
These datasets consist of images captured under different lighting situations and adverse conditions, each with manually created semantic segmentation paired with a reference image taken under favorable weather conditions.
This has paved the way for weakly-supervised domain adaptation methods that leverage the information in this association.

Sakaridis \etal~\cite{sakaridis2022} gradually adapt from the daytime to the nighttime domain using images from different lighting situations.
They extract information from the reference-to-target image pairing and use artificially stylized annotated images.
Bruggemann \etal~\cite{Bruggemann2023} introduced an uncertainty-aware ALIGN-module to improve the alignment between reference and target domains, omitting the usage of stylized images and the gradual adaptation procedure.

In another work by Bruggemann \etal~\cite{Bruggemann2023CMA}, a contrastive loss formulation in the embedding space is used to learn weather-condition-robust features.
This procedure requires maintaining a list of negative samples for each positive sample, which can complicate implementation and impede training if the negatives are too close to the positives.

\subsection{Self-supervised Learning}
\label{subsec:ssl}

Self-supervised learning methods have been developed to leverage the intrinsic structure of unlabeled data.
These methods typically generate synthetic augmentations of input images, training networks to produce feature representations that are invariant to such changes.
However, without careful design, networks may collapse to a trivial solution where they output constant features that do not depend on the data.

To mitigate this, contrastive-based methods, such as those by Chen \etal~\cite{chen2020simple} and Misra \etal~\cite{misra2020pirl}, have been introduced.
These methods enforce a closer similarity between features of augmented images from the same source (positive pairs) than those from different sources (negative pairs). 
The selection of negative samples is crucial in these approaches to prevent the network from settling into degenerate solutions.

Clustering-based approaches~\cite{caron2020unsupervised,asano2020self}, group images into clusters, aiming to ignore any augmentations.
These methods must carefully manage the clustering process to avoid all embeddings converging into a single cluster, another form of the trivial solution.

The Barlow Twins method, introduced by Zbontar \etal~\cite{zbontar2021}, employs redundancy reduction techniques, notably avoiding the use of negative pairs, asymmetric updates, stopped gradients, or non-differentiable operators.
It circumvents the trivial solution through a loss function that encourages the cross-correlation matrix of batch-normalized embeddings of augmented image pairs to approximate the identity matrix.

Building on this, Bardes \etal~\cite{bardes2022} presented VICReg, which incorporates two separate regularization terms applied to each embedding and a third term to ensure the embeddings remain similar.
This model introduces additional hyperparameters due to the need to balance these terms.
However, since VICReg did not demonstrate significant improvements over the Barlow Twins' approach, we have chosen to employ the latter in our method.

\subsection{Barlow Twins for Semantic Segmentation}

The Barlow Twins methodology has been adopted in the segmentation of medical imagery with notable success, as reported by Siddiquee \etal~\cite{Siddiquee2021} and Punn \etal~\cite{Punn2022}.
In contrast to our methodology, these approaches depend on conventional augmentation techniques (e.g., cropping, resizing, blurring), outlined in the seminal work by Zbontar~\etal~\cite{zbontar2021}.

The work of Bhagwatkar \etal~in~\cite{Bhagwatkar2022} exploits the Barlow Twins loss formulation~\cite{zbontar2021} to train an encoder network robust against ``domain transformations'' and assumes ``corresponding images from different domains as different views of the same abstract representation''.
Additionally, they jointly train a segmentation head for one domain, utilizing the output of the encoder.
However, they try to adapt from synthetic to real data, but it remains unclear how to select corresponding images in this setting.
As a result, unlike our method, the images they use lack closely related semantic content.
We posit that this could hinder the encoder training, as it must reconcile the creation of similar embeddings for dissimilar image contents.
While this may enhance robustness to domain shifts, it simultaneously compels the encoder to reconcile divergent image contents, presenting conflicting training goals.

In contrast, our work utilizes natural, paired images of similar locales captured in both clear and adverse environmental conditions.
Furthermore, we implement sophisticated techniques to manage dynamic content and reconcile misaligned perspectives, thus addressing issues beyond those tackled by previous methods.

\section{Method}

This section outlines our methodology, which exploits the Barlow Twins loss to fortify semantic segmentation networks against domain variations.

\subsection{Brief Introduction to Barlow Twins}
\label{subsec:BT}

The Barlow Twins loss was introduced by Zbontar \etal~\cite{zbontar2021} and originates from the domain of self-supervised learning.
In their approach, images from the training dataset are passed through two randomized augmentation pipelines, to create two varied versions ($\mathbf{X}_A$ and $\mathbf{X}_B$) of the same original input image.
They are then fed into two Siamese encoder networks which yields two separate \textit{representations}.
Subsequently, these representations are condensed via an average pooling layer and further processed by an attached fully-connected projection network.
For a batch of augmented images, this results in two \textit{embedding} vectors, $\mathbf{Z}_A$ and $\mathbf{Z}_B \in M_{b \times p}(\mathbb{R})$ with $b$ being the batch size and $p$ the dimension of the embedding vector.
The key idea is to normalize each dimension of the embedding vectors across its batch dimension and to use the resulting normalized vectors to calculate the cross-correlation matrix $\mathbf{C} \in M_{p \times p}(\mathbb{R})$.
Correlation between the vectors is then enforced by the first part of the proposed loss function

\begin{equation}
\label{eq:loss_bt}
    \mathcal{L_{BT}} = \sum_{i}(1 - \mathbf{C}_{ii})^2 + \lambda \sum_{i} \sum_{j \ne i} {\mathbf{C}_{ij}}^2
\end{equation}

that enforces values of one on the diagonal and thus invariance against the applied augmentations.
The second part urges zeros everywhere else, therefore reducing the information redundancy, while $\lambda \in \mathbb{R}$ equilibrates both parts.

The main advantage arises from the absence of a requirement for negative samples, as in contrastive losses.

\subsection{Robust Semantic Segmentation through Barlow Twins}
\label{subsec:bt_seg}

\begin{figure*}[t]
     \centering
     \includegraphics[width=0.9\textwidth]{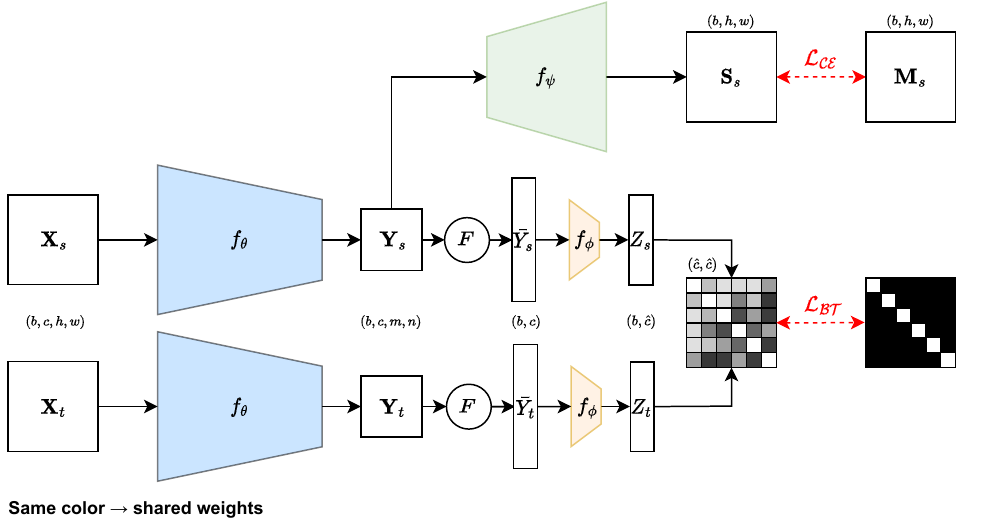}
     \vspace{-0.3cm}
     \caption{Schematic overview of our proposed semi-supervised semantic segmentation architecture, inspired by \cite{zbontar2021}. But instead of creating paired images with augmentations from one base image, we use images of the same location captured under different environmental conditions. We also train the semantic segmentation head directly, using the Barlow Twins loss as a regularization method, rather than as a pre-training strategy. To mitigate the effects of moving objects, we introduced an additional guided pooling operation $F$. During inference, the network utilizes only the upper path to generate the segmentation mask.}
     \label{fig:network_flow}
\end{figure*}

Our network architecture, shown in \cref{fig:network_flow}, integrates the Barlow Twins objective into a typical encoder ($f_\theta$)-decoder ($f_\psi$) structure for semantic segmentation.
We use two images of the same location under different conditions (e.g., seasonal or weather changes) as natural augmentations within the Barlow Twins framework. 
We assume these images are accurately registered and semantically aligned, with strategies for non-aligned content explained in \crefrange{subsec:filteringMobile}{subsec:confidence_averging}.

In domain adaptation terms, the image with favorable conditions is the source image $\mathbf{X_s}$, and the one with adverse conditions is the target image $\mathbf{X_t}$. 
Both images, with dimensions $(b,c,h,w)$, pass through the encoder $f_{\theta}$, producing feature maps $\mathbf{Y_s}$ and $\mathbf{Y_t}$ with reduced dimensions $(b,d,m,n)$. 
These maps are pooled by $F$ into $(b,d,1,1)$, yielding $\mathbf{\bar{Y}}_s$ and $\mathbf{\bar{Y}}_t$.

Following the Barlow Twins protocol (\cref{subsec:BT}), the pooled features are fed into a projection network $f_{\phi}$, producing embeddings $\mathbf{Z_s}$ and $\mathbf{Z_t} \in M_{b \times \hat{c}}(\mathbb{R})$. 
These embeddings are normalized and cross-correlated to form matrix $\mathbf{C} \in M_{\hat{c} \times \hat{c}}(\mathbb{R})$, establishing the first objective $\mathcal{L_{BT}}$ (\cref{eq:loss_bt}) for domain-robust representations.

The source path representations are processed by the decoder $f_\psi$, resulting in the semantic segmentation map $\mathbf{S_s}$, which is resized to the original input size and evaluated using the Cross Entropy loss $\mathcal{L_{CE}}$.
The combined objectives (\cref{eq:overall_loss}) guide the network to robust, efficient representations for semantic segmentation.

\begin{equation}
\label{eq:overall_loss}
\mathcal{L} = \mathcal{L_{CE}} + \alpha \mathcal{L_{BT}}
\end{equation}

Our approach exclusively uses annotated reference images with favorable conditions, avoiding the need to label adverse images.
This is advantageous as clear weather images are easier to annotate.
While assuming domain shifts are the primary variations, the following sections address strategies for spatial misalignments common in real-world conditions.

\begin{figure*}[t]
    \centering

    \begin{subfigure}[b]{0.24\textwidth}
            \includegraphics[width=\textwidth]{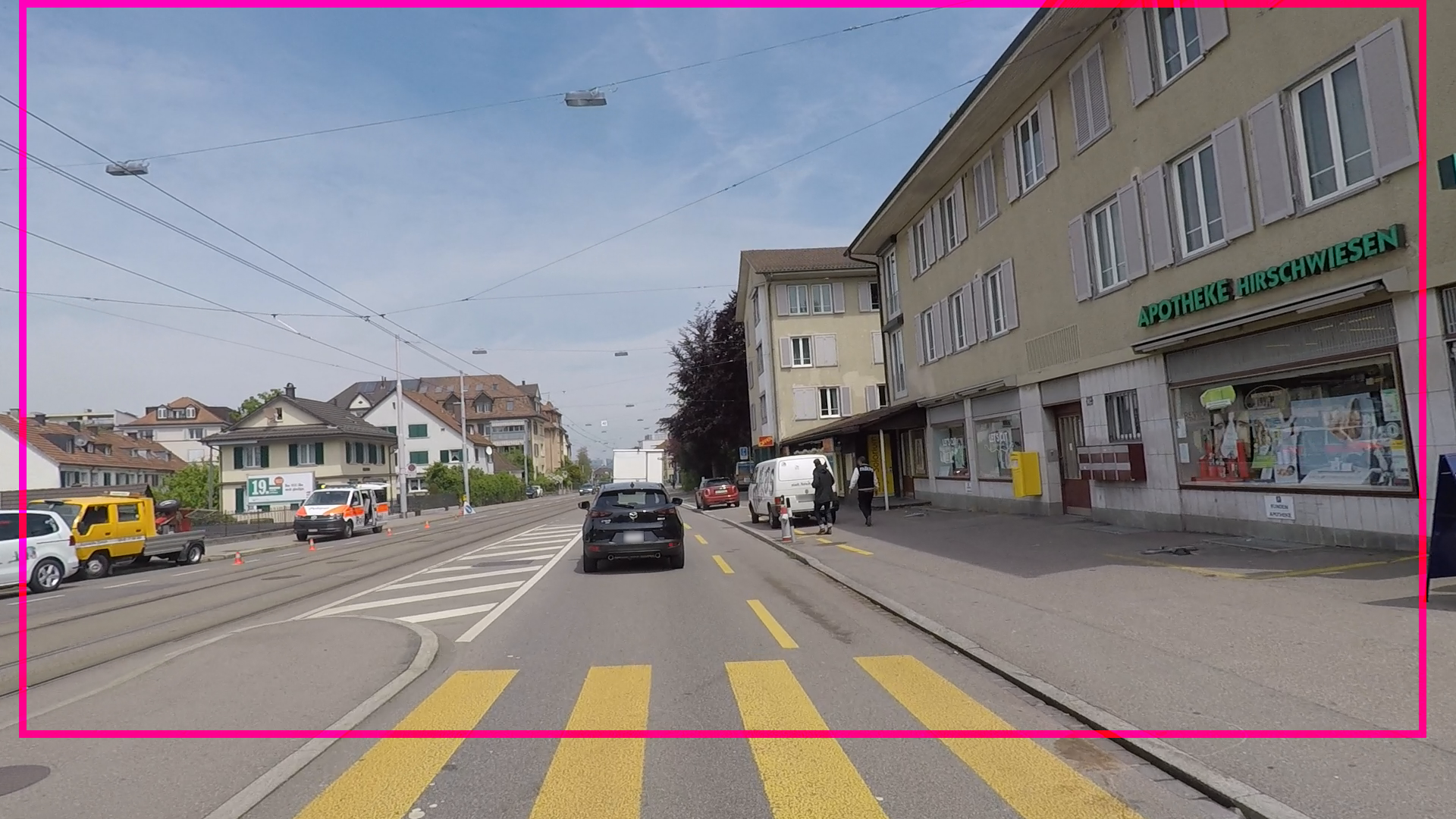}
            \caption{Source}
    \end{subfigure}
    \begin{subfigure}[b]{0.24\textwidth}
            \includegraphics[width=\textwidth]{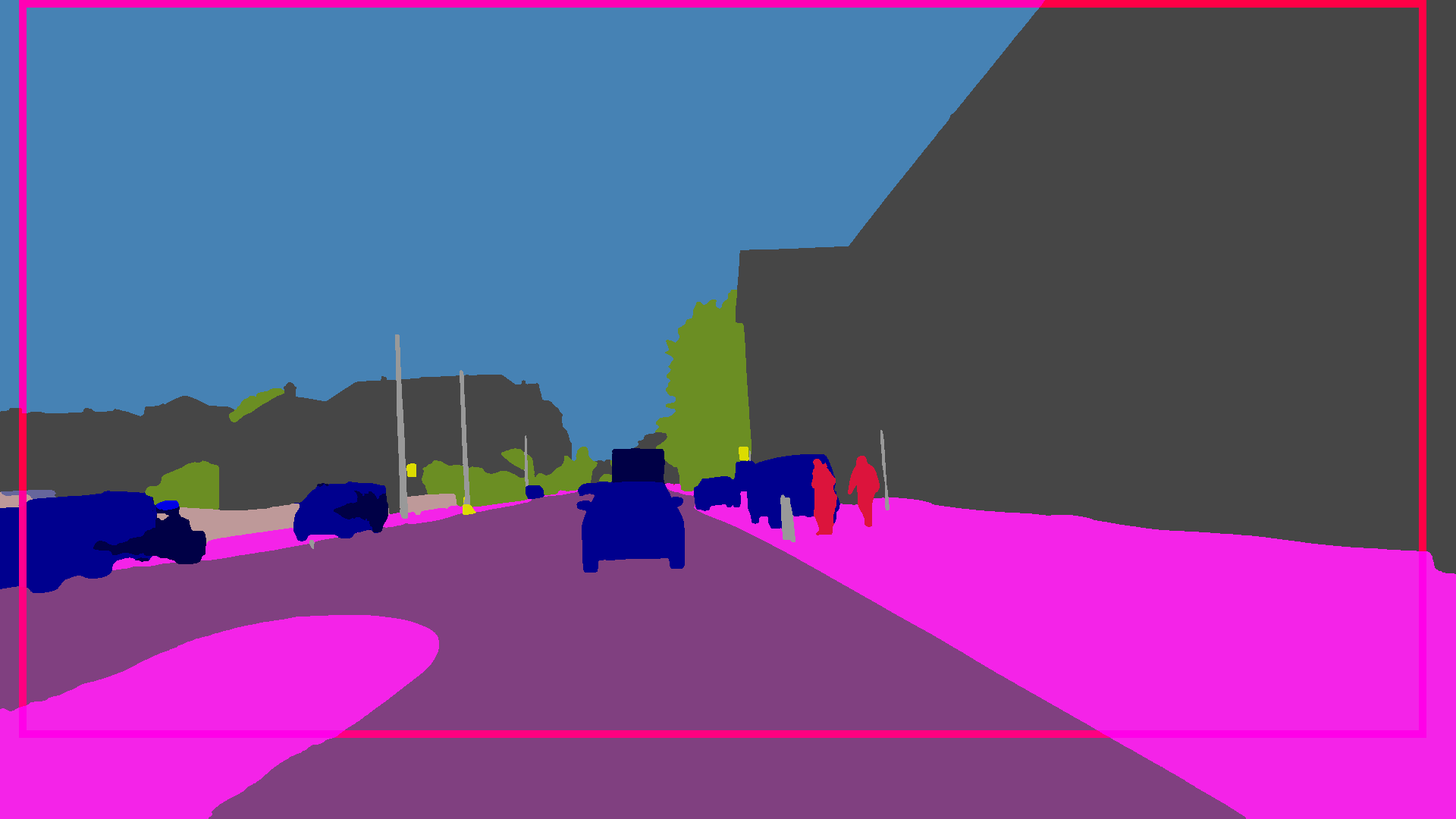}
            \caption{Source GT}
    \end{subfigure}
    \begin{subfigure}[b]{0.24\textwidth}
            \includegraphics[width=\textwidth]{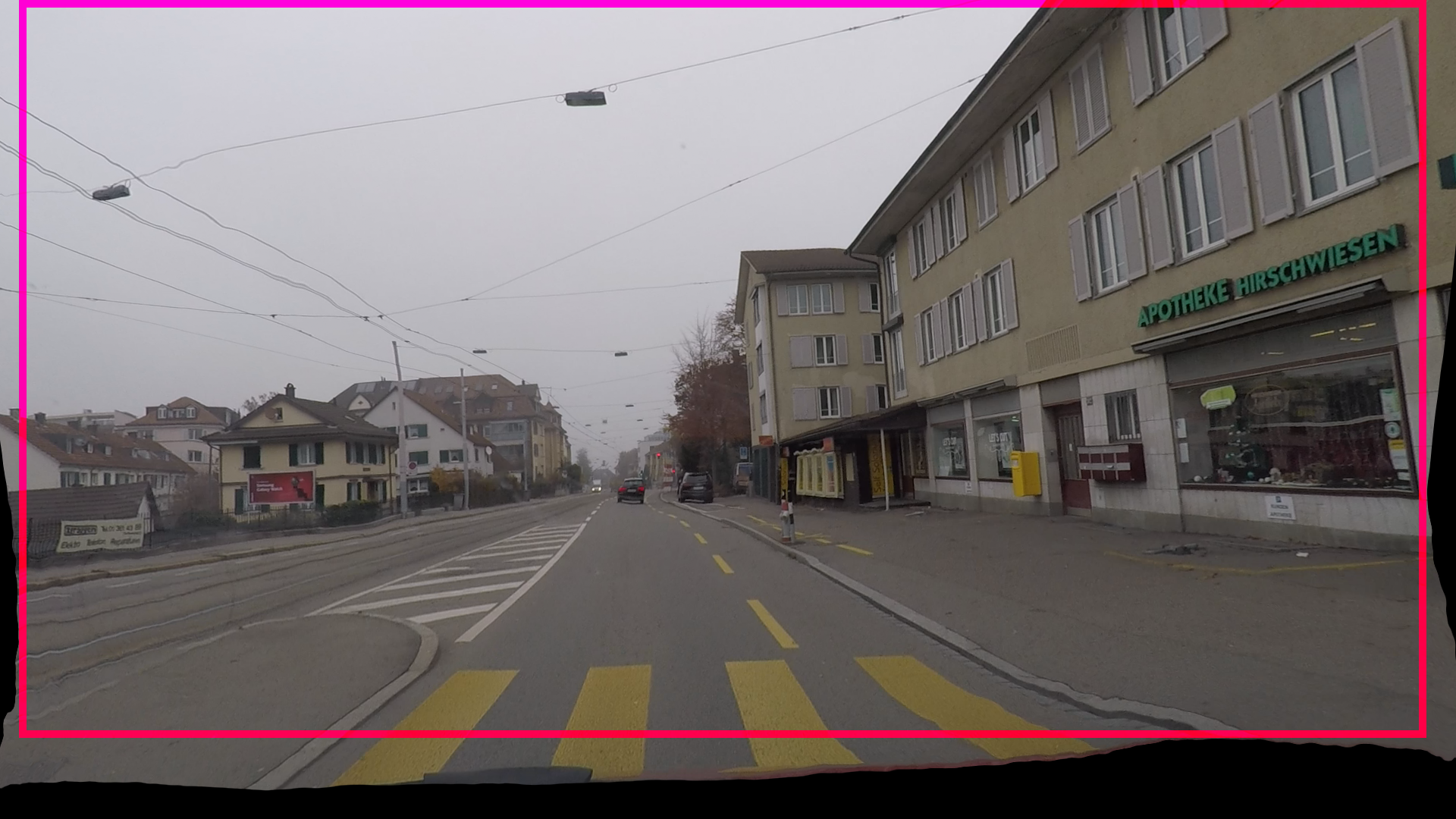}
            \caption{Warped target}
    \end{subfigure}
    \begin{subfigure}[b]{0.24\textwidth}
            \includegraphics[width=\textwidth]{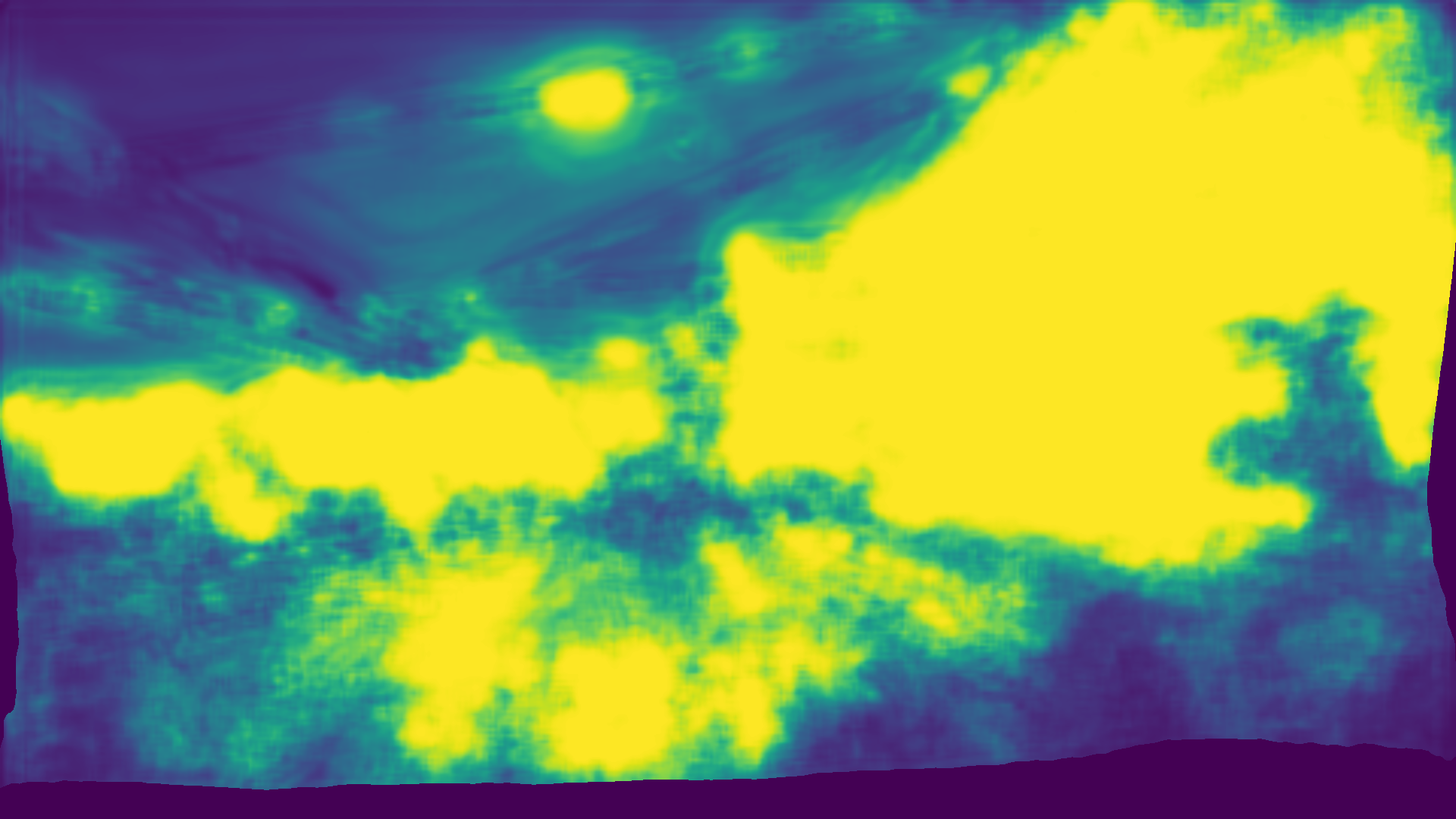}
            \caption{Warp Confidence}
            \label{fig:align_conf}
    \end{subfigure}
    
    \caption{Sample image from the ACDC dataset with the image from the target domain warped to the source domain to align the semantics. The confidence map allows a prediction of the warp quality and is used as described in \cref{subsec:confidence_averging}. The \textit{red rectangles} denote the largest intrinsic rectangles containing only valid image content.}
    \label{fig:align}

\end{figure*}

\subsection{Pre-alignment of Source and Target}
\label{subsec:prealign}

To reduce the influence of misalignments between the semantic content in the image pairs caused by slightly different viewpoints, we pre-align the target (adverse condition) to the source (normal condition) image, utilizing the \textit{UAWarpC}-module~\cite{Bruggemann2023}, pre-trained in a self-supervised manner on MegaDepth~\cite{Li2018}.
The method returns a warp map aligning the two images, with zero indicating invalid regions (see \cref{fig:align}).
The latter is used to find the largest intrinsic rectangle (marked red in \cref{fig:align}) containing the valid warped image region, utilizing the algorithm\footnote{Implementation from https://github.com/OpenStitching/lir} introduced by Marzeh~\etal~\cite{Marzeh2019}.
Source, target, and the source segmentation are cropped accordingly.

\subsection{Enhancing Robustness Amidst Moving Objects}
\label{subsec:filteringMobile}
By default, the pooling operation $F$ calculates a single average value for each channel of the whole feature map $\mathbf{Y}$.
While this is already a fairly strong smoothing operation, moving objects, e.g.~cars, could disrupt the learning of weather-induced differences in the final embeddings $\mathbf{Z}$.
To this end, we extend the default average pooling $F$ with guidance from the segmentation heads.
We generate a binary mask $\mathbf{B} \in M_{b \times m \times n}(\mathbb{R})$ from $\mathbf{S}$ with

\begin{equation}
    \mathbf{B}_{i,j} =
    \begin{cases}
      0 & \text{if $i,j$ was labeled as a moving object}\\
      1 & \text{otherwise}
    \end{cases},  
\end{equation}

to filter areas identified by the segmentation head ($f_\psi$) as moving objects (person, rider, car, truck, bus, train, motorcycle and bicycle).
Consequently, the average pooling with the exclusion of areas containing moving objects can be formulated as

\begin{equation}
    \bar{\mathbf{Y}} = \frac{\sum ( \mathbf{Y} \odot \mathbf{B} ) }{\sum{ (\mathbf{B}) }+ \epsilon},
\end{equation}

with $\epsilon$ being some small constant for numerical stability, $\sum( \cdot )$ the sum over all elements in a matrix and $\odot$ indicating the element-wise product between two matrices.

\subsection{Confidence-weighted averaging}
\label{subsec:confidence_averging}

The \textit{UAWarpC} module outputs a confidence estimation $\mathbf{P}_W$ , depicted in \cref{fig:align_conf}, with values within the interval [0,1].
This confidence map assigns higher confidence scores to well-matched image regions, such as facades, poles, and trees, and lower scores to areas with less distinctive textures, like the sky or roads.
Additionally, moving objects are assigned lower confidence unless, by coincidence, they appear in the same image region.
These confidence values are then employed to compute a weighted average for the feature maps $\mathbf{Y}$.

\section{Experiments}
\label{sec:exp}

In this section, we describe the datasets (\cref{subsec:datasets}) and implementation details (\cref{subsec:implementationdetails}). We compare our method quantitatively and qualitatively to the state of the art in weakly-supervised domain adaptation (\cref{subsec:sota}), evaluate design choices (\cref{subsec:ablation}), and assess robustness and generalization (\cref{subsec:acg_benchmark}). Additional experiments and more qualitative results can be found in the supplementary material.

\subsection{Datasets}
\label{subsec:datasets}

\paragraph{ACDC} Sakaridis~\etal introduced the ACDC~(Adverse Conditions Dataset with Correspondences), specifically designed for semantic segmentation in autonomous driving under adverse weather conditions~\cite{Sakaridis2021}.
For each weather condition (night, rain, fog, snow), 400 training, 100 validation (106 for night) and 500 testing images are available.
Every image in the dataset is coupled with a counterpart showing the same scene under clear weather conditions, approximately aligned using GPS data.
Originally, the dataset only contains segmentation masks for the adverse conditions, which we employ solely for testing purposes.
To generate ground truth labels for the clear condition images, we leveraged a state-of-the-art model \cite{Tao2020}, trained on Cityscapes, to generate semantic segmentations for the reference images.

To optimize training, we refined the dataset by filtering out unsuitable samples, when used in conjunction with our pre-alignment process (see \cref{subsec:prealign}).
A detailed description of the filtering process is provided in the supplementary material.
It resulted in reduced sample counts of 346, 370, 277, and 319 for night, rain, fog, and snow conditions, respectively.
The details of the excluded samples are available in the code accompanying our publication.

\paragraph{ACG Benchmark} Bruggemann~\etal~\cite{Bruggemann2023CMA} introduced the ACG (adverse-condition generalization) benchmark dataset to rigorously assess the generalization capabilities of models following domain adaptation.
The dataset was created through careful selection of images from WildDash2~\cite{Zendel_2022_CVPR}, Foggy Zurich~\cite{SDHV18,Dai2019CurriculumMA}, Foggy Driving~\cite{SDV18} and BDD100K~\cite{Yu2018BDD100KAD} and contains 121 images with fog, 225 with rain, 276 with snow, and 300 taken during nighttime conditions.
In contrast to the ACDC dataset, this dataset includes images with various combined adverse conditions, such as night-time rain, which makes it even more challenging.
We exclusively utilize this dataset for testing purposes.

\subsection{Implementation Details}
\label{subsec:implementationdetails}

As proposed by Tsai \etal~\cite{Tsai2021}, we set $\lambda$ in \cref{eq:loss_bt}  depending on the fraction of on-diagonal ($=n$) and off-diagonal ($=n(n-1)$) elements following

\begin{equation}
\label{eq:lambda}
    \frac{n}{n(n-1)} \approx \frac{1}{n} = \lambda,
\end{equation}

in order to avoid the necessity of tuning it.

We evaluated BTSeg in conjunction with SegFormer~\cite{xie2021}, a widely adopted architecture in weakly-supervised semantic segmentation literature.
However, we want to emphasize that BTSeg is highly adaptable and can be integrate into various semantic segmentation architectures equipped with an encoder backbone and detection head.
Please refer to the supplementary, for the evaluation of BTSeg in conjunction with DeeplabV3~\cite{chen2017}.
Following Bruggemann~\etal~\cite{Bruggemann2023CMA}, we resized and concatenated transformer block outputs of SegFormer to match the highest resolution block, resulting in 1024-dimensional feature maps $\mathbf{Y}$.

Our projection network comprises two layers with \gls{bn} and ReLU activation, reducing input dimensions from 1024 to 512, and then to 256. 
This contrasts with the original model's three-layer, 8192-unit projection network, which we found excessively large without added benefit.
A comparison of different configurations can be found in our supplementary.

We strictly used complete batches during training to accommodate the batch-normalization layer required for the Barlow Twins loss.
Consistent with common practices in the literature, we conducted pretraining of SegFormer on the Cityscapes~\cite{Cordts2016} dataset.

Due to the substantial memory requirements of the SegFormer architecture, Bruggemann~\etal~\cite{Bruggemann2023CMA} implemented gradient accumulation while still adhering to a batch size of 2.
However, the Barlow Twins loss, incorporating batch normalization (\gls{bn}), demands higher batch sizes and is incompatible with gradient accumulation. 
Nonetheless, as the \gls{bn} in the Barlow Twins loss doesn't involve learnable parameters and is solely used for normalization across the batch dimension, we opt to process images individually.
The resulting embeddings $\mathbf{Z}$ are stored in a cache, allowing us to effectively achieve a batch size of 32 for both the Barlow Twins loss and the overall network, while still employing gradient accumulation.
To ensure that the larger batch size does not give BTSeg an undue advantage, we conducted an additional experiment to rule out its influence on our closest rival, CMA.
The details of this experiment can be found in the supplementary material.

We trained for 10000 steps on one single GPUs using automatic mixed precision.
Data augmentation included random (but consistent for both, $\mathbf{X_s}$ and $\mathbf{X_t}$  horizontal flipping and cropping to 768x768 (the smallest intrinsic rectangle, as detailed in \cref{subsec:datasets}, ACDC paragraph), followed by resizing to 1080x1080 to maintain the original resolution.

In our combined objective function (\cref{eq:overall_loss}), we set $\alpha$ to 0.1 to balance the loss magnitudes.
We employed AdamW optimizer with a 0.01 weight decay, 1500 steps of linear warmup, followed by a linear decay.
For the first 2500 steps, we stopped the gradient flow from the projection head to the encoder to avoid noisy updates.
We used different learning rates for encoder (\num{1.6e-4}), decoder (\num{1.6e-5}) and the projection network (\num{1.6e-3}).

\subsection{Comparison to SOTA in Weakly-supervised Domain Adaptation}
\label{subsec:sota}

\begin{table}
\caption{Comparison to the state of the art in semi-supervised domain adaptation from Cityscapes $\rightarrow$ ACDC, with the results reported on the official ACDC test set.}
\label{tab:sota}
\smallskip%
\centering%
\ra{1.3}%
\def\mycolw{\textnormal{\textsc{alig}} }
\resizebox{\textwidth}{!}{
\begin{tabular}{@{}lcccccccccccccccccccc@{}}\toprule
\multirow{2}{*}{\textbf{Method}} &  \multicolumn{20}{c}{\textbf{IoU}$\uparrow$} \\
\cmidrule{2-21} & \rotatebox[origin=c]{90}{\textbf{mean}} & \rotatebox[origin=c]{90}{road} & \rotatebox[origin=c]{90}{sidew.} & \rotatebox[origin=c]{90}{build.} & \rotatebox[origin=c]{90}{wall} & \rotatebox[origin=c]{90}{fence} & \rotatebox[origin=c]{90}{pole} & \rotatebox[origin=c]{90}{light} & \rotatebox[origin=c]{90}{sign} & \rotatebox[origin=c]{90}{veget.} & \rotatebox[origin=c]{90}{terrain} & \rotatebox[origin=c]{90}{sky} & \rotatebox[origin=c]{90}{person} & \rotatebox[origin=c]{90}{rider} & \rotatebox[origin=c]{90}{car} & \rotatebox[origin=c]{90}{truck} & \rotatebox[origin=c]{90}{bus} & \rotatebox[origin=c]{90}{train} & \rotatebox[origin=c]{90}{motorc.} & \rotatebox[origin=c]{90}{bicycle} \\ \midrule
SegFormer\textsubscript{512}~\cite{xie2021} & \textover[c]{50.68}{\mycolw} & 76.4 & 42.0 & 70.8 & 32.0 & 29.6 & 40.3 & 47.3 & 49.2 & 68.1 & 36.9 & 76.5 & 39.8 & 22.6 & 77.0 & 49.4 & 60.7 & 65.3 & 33.5 & 45.1\\
SegFormer\textsubscript{1024}~\cite{xie2021} & 59.34 & 85.7 & 51.0 & 76.6 & 36.4 & 37.1 & 45.2 & 55.7 & 57.5 & 77.7 & 51.9 & 84.2 & 60.2 & 34.7 & 83.0 & 61.6 & 65.5 & 73.4 & 37.9 & 52.4\\
\midrule
MGCDA \cite{sakaridis2022} & 48.65 & 73.1 & 29.0 & 70.0 & 19.2 & 26.4 & 36.7 & 52.7 & 53.2 & 75.0 & 31.7 & 84.8 & 50.6 & 25.7 & 77.7 & 43.1 & 45.8 & 55.0 & 32.9 & 41.7\\
Refign \cite{Bruggemann2023} & 65.51 & 89.5 & 63.4 & 87.3 & 43.6 & 34.3 & 52.3 & 63.2 & 61.4 & 87.0 & 58.5 & 95.7 & 62.1 & 39.3 & 84.1 & 65.7 & 71.3 & 85.4 & 47.9 & 52.8\\
CMA \cite{Bruggemann2023CMA}  & 69.12 & \textbf{94.0} & \textbf{75.2} & \textbf{88.6} & 50.5 & 45.5 & 54.9 & 65.7 & 64.2 & \textbf{87.1} & 61.3 & 95.2 & 67.0 & 45.2 & 86.2 & 68.6 & 76.6 & \textbf{84.0} & 43.3 & \textbf{60.5}\\
\midrule
BTSeg (Ours) & \textbf{70.45} & 92.9 & 73.5 & 88.3 & \textbf{53.6} & \textbf{47.2} & \textbf{56.2} & \textbf{67.9} & \textbf{67.3} & 86.9 & \textbf{61.5} & \textbf{95.3} & \textbf{67.2} & \textbf{46.8} & \textbf{89.1} & \textbf{73.1} & \textbf{81.6} & 83.1 & \textbf{49.5} & 57.5\\
\bottomrule
\end{tabular}}
\end{table}

In \cref{tab:sota}, we compare BTSeg against the current state of the art for weakly-supervised domain adaptation, transitioning from Cityscapes to ACDC.
All methods, with the exception of MGCDA~\cite{sakaridis2022}, employ Mix Transformer (MiT-B5)~\cite{xie2021} as an encoder.
The evaluation was conducted on the official ACDC test set, with the images processed directly with two overlappping 1080x1080 patches.
The first two lines contain the baseline, a SegFormer network trained solely on Cityscapes in a fully-supervised fashion, as described in~\cite{xie2021}.
We conducted training on two variations: one utilizing the complete resolution of Cityscapes in conjunction with 1024x1024 pixel patches, and another with images and patches scaled by 0.5, to enable a fair comparison, as MGCDA also utilized input images with half resolution.
Based on a Transformer architecture, the baseline already shows good generalization properties, and training with full resolution already improves the mean IoU (Intersection over Union) by \SI{8.66}{\pp}.
Our method outperforms Refign~\cite{Bruggemann2023} and CMA~\cite{Bruggemann2023CMA}, while at the same time being less complex, as the Barlow Twins loss avoids the necessity of maintaining a list of negative samples, which are needed e.g.~for the contrastive loss in CMA.
Similarly to CMA, our method shows a good performance for moving objects.
This is remarkable, as these classes are especially challenging to our proposed method, as they contradict our assumption of shared content between reference and target image.
We hypothesize that this is due to the rather coarse regularization and the reduction of the $(c, m, n)$-dimensional feature map to a $(c, 1, 1)$-dimensional feature vector.
In summary, our method establishes a new state-of-the-art for weakly supervised domain adaptation.

\begin{figure}[ht]
    \centering
    \begin{tabular}{ccccc}
        \raisebox{0.5cm}{\begin{sideways}\tiny Input\end{sideways}} &
        \includegraphics[width=0.23\textwidth]{  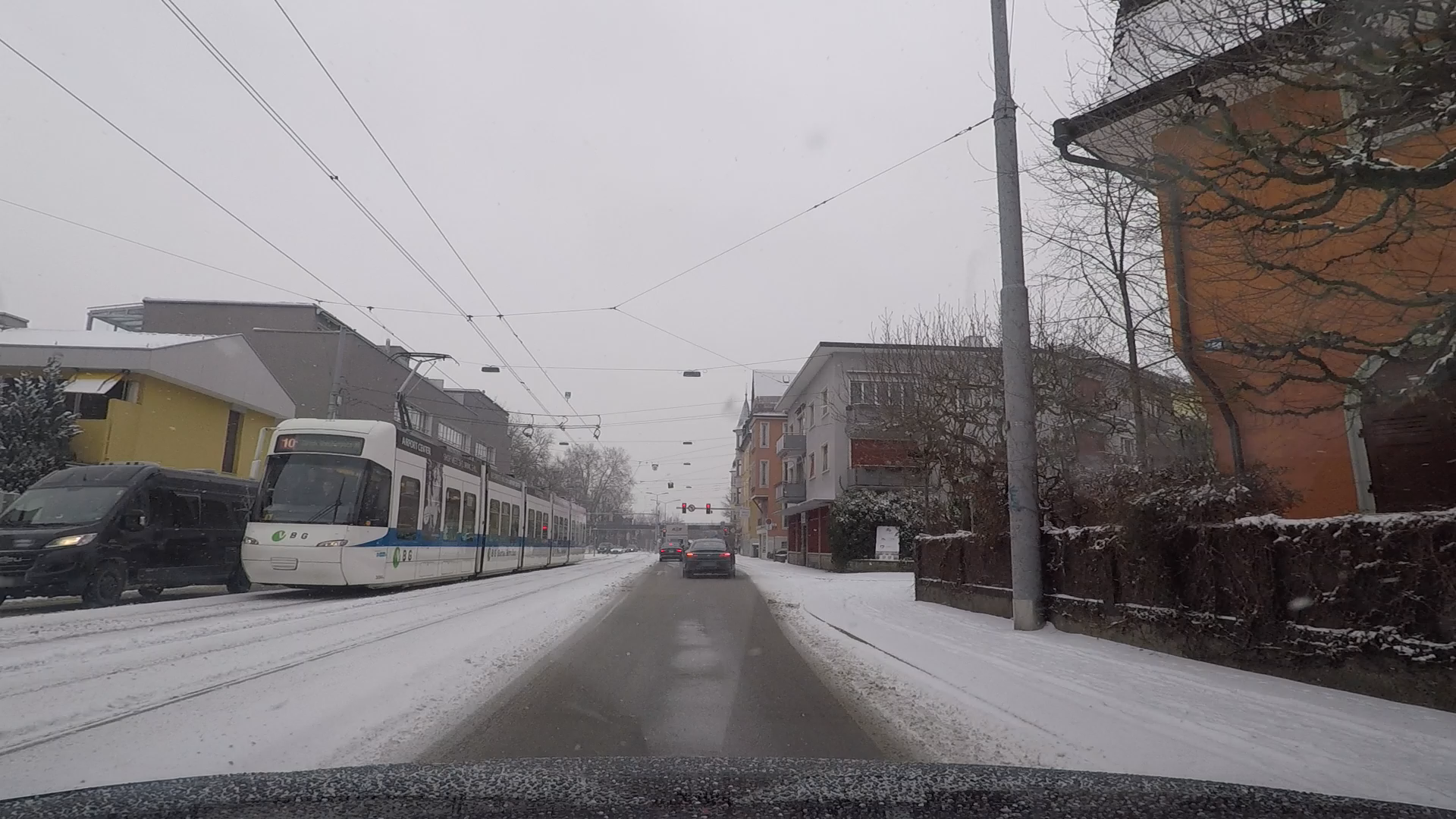} &
        \includegraphics[width=0.23\textwidth]{  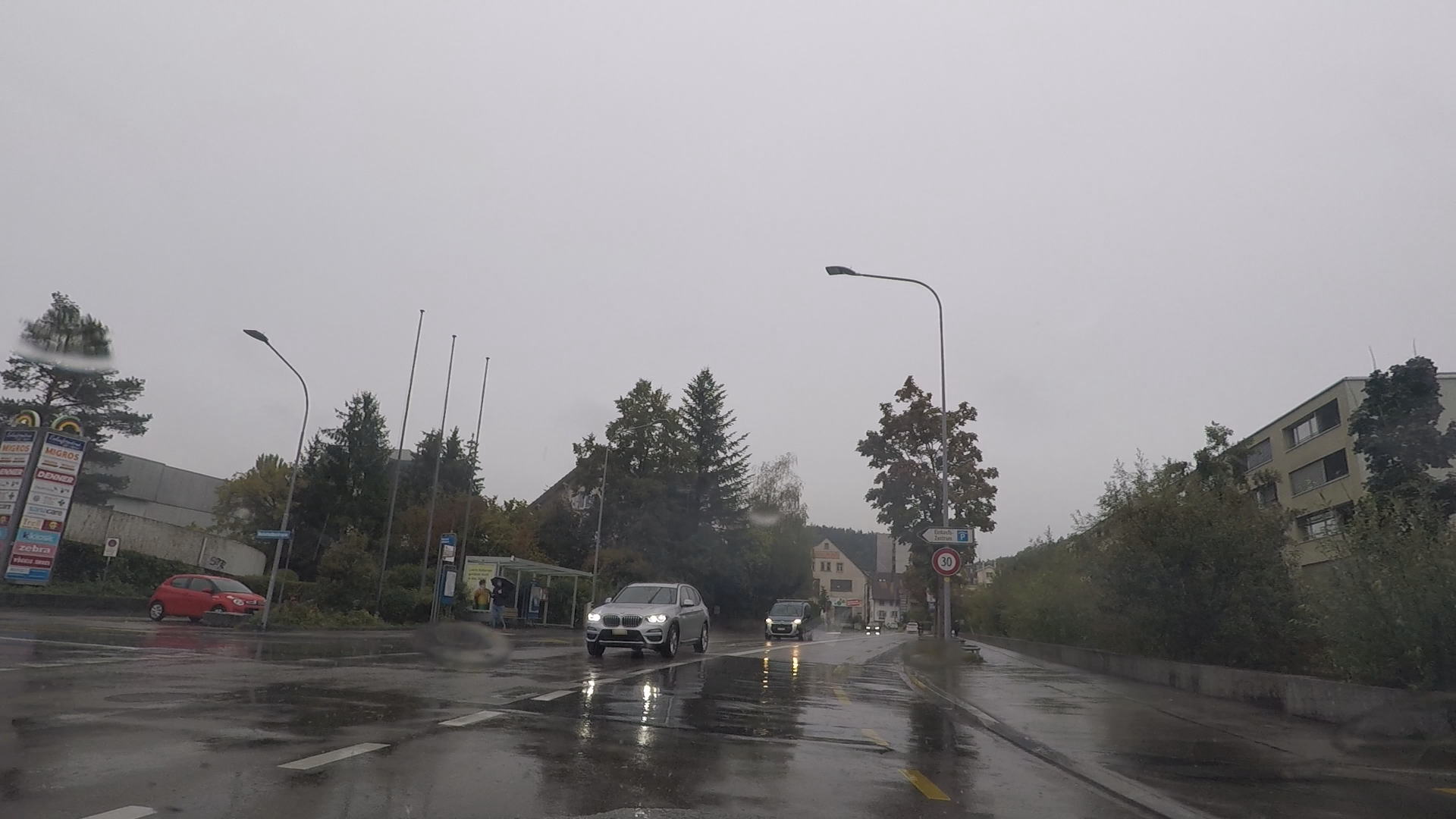} &
        \includegraphics[width=0.23\textwidth]{  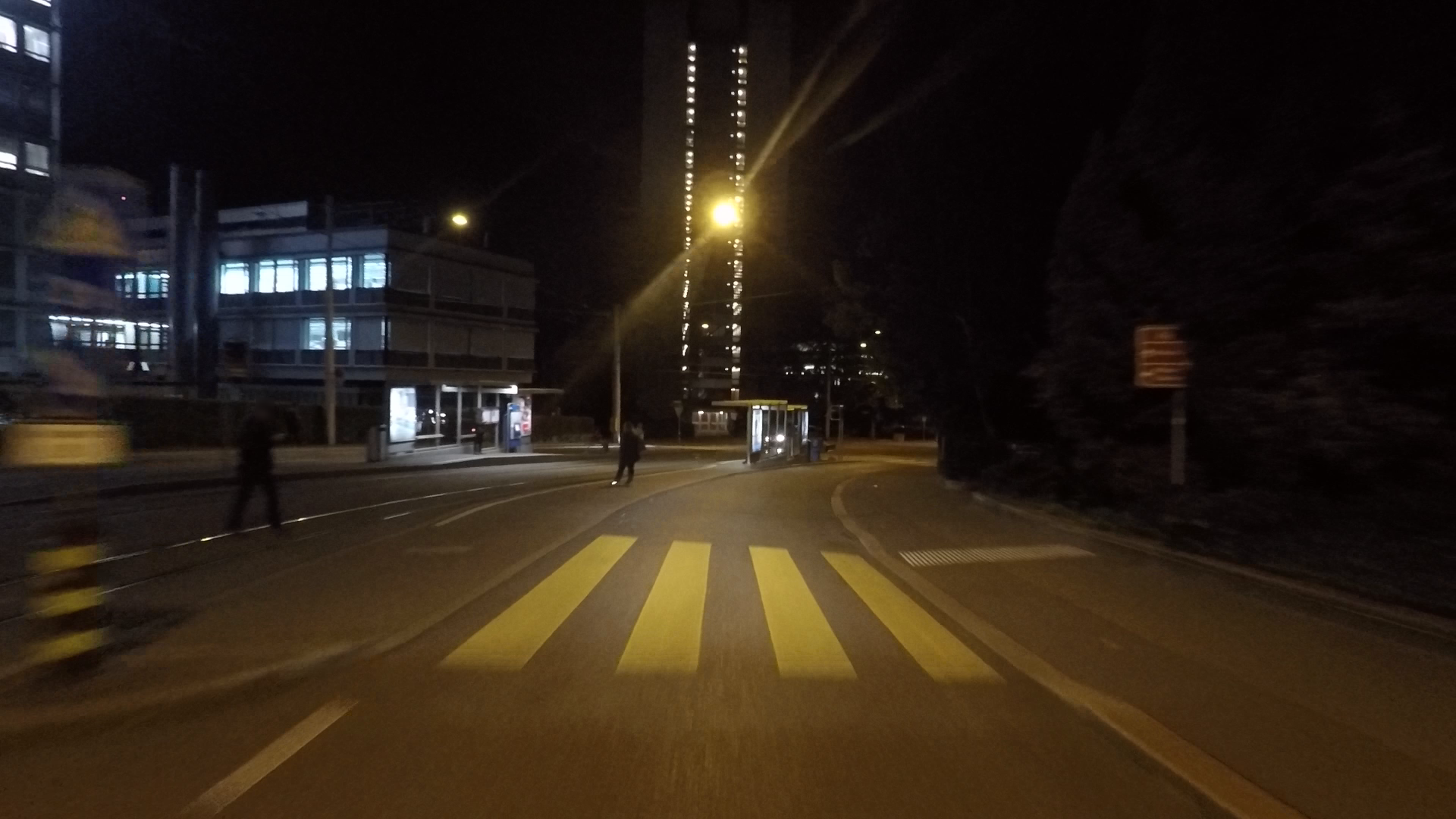} &
        \includegraphics[width=0.23\textwidth]{  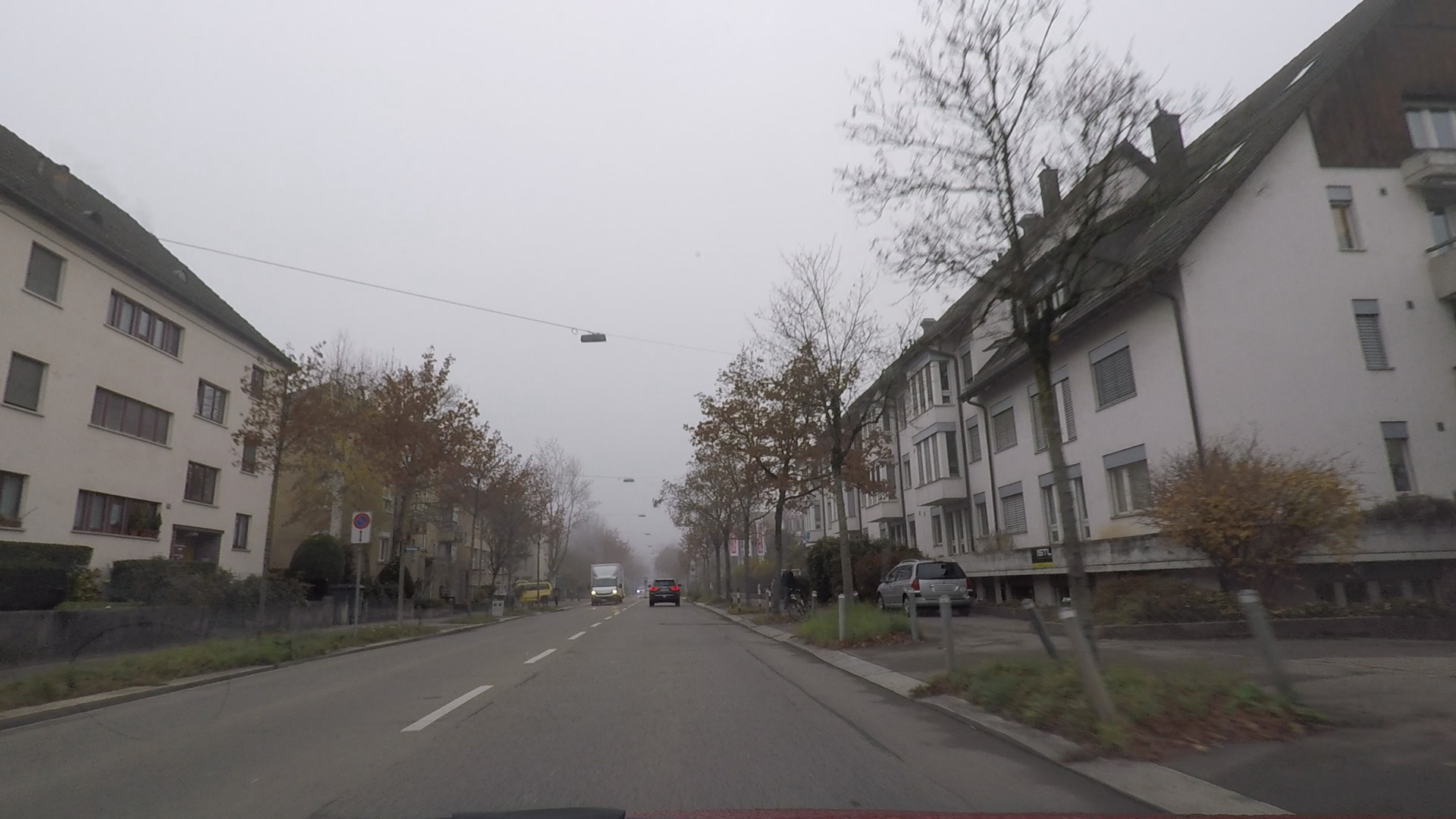} \\

        \raisebox{0.6cm}{\begin{sideways}\tiny GT\end{sideways}} &
        \includegraphics[width=0.23\textwidth]{  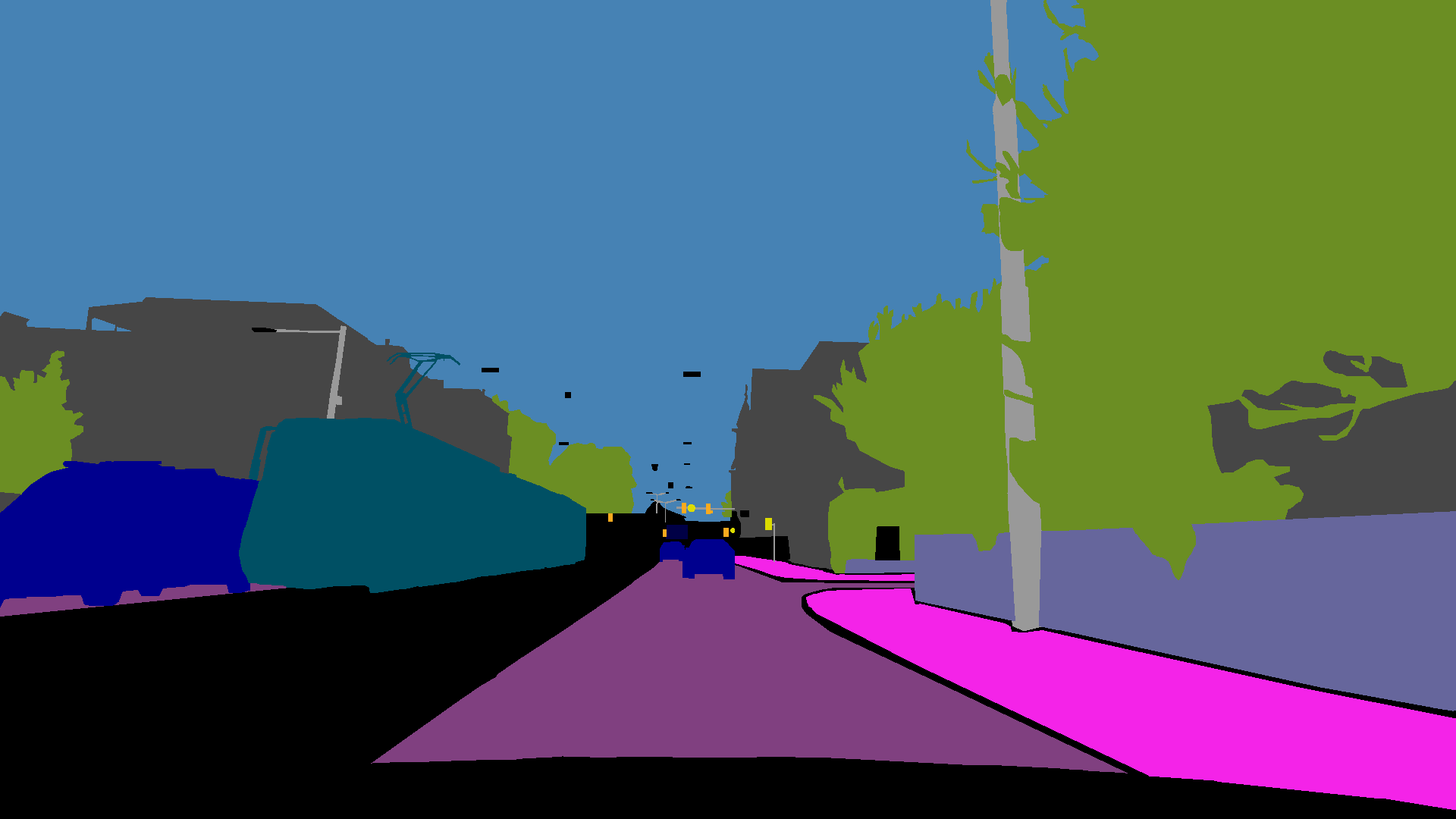} &
        \includegraphics[width=0.23\textwidth]{  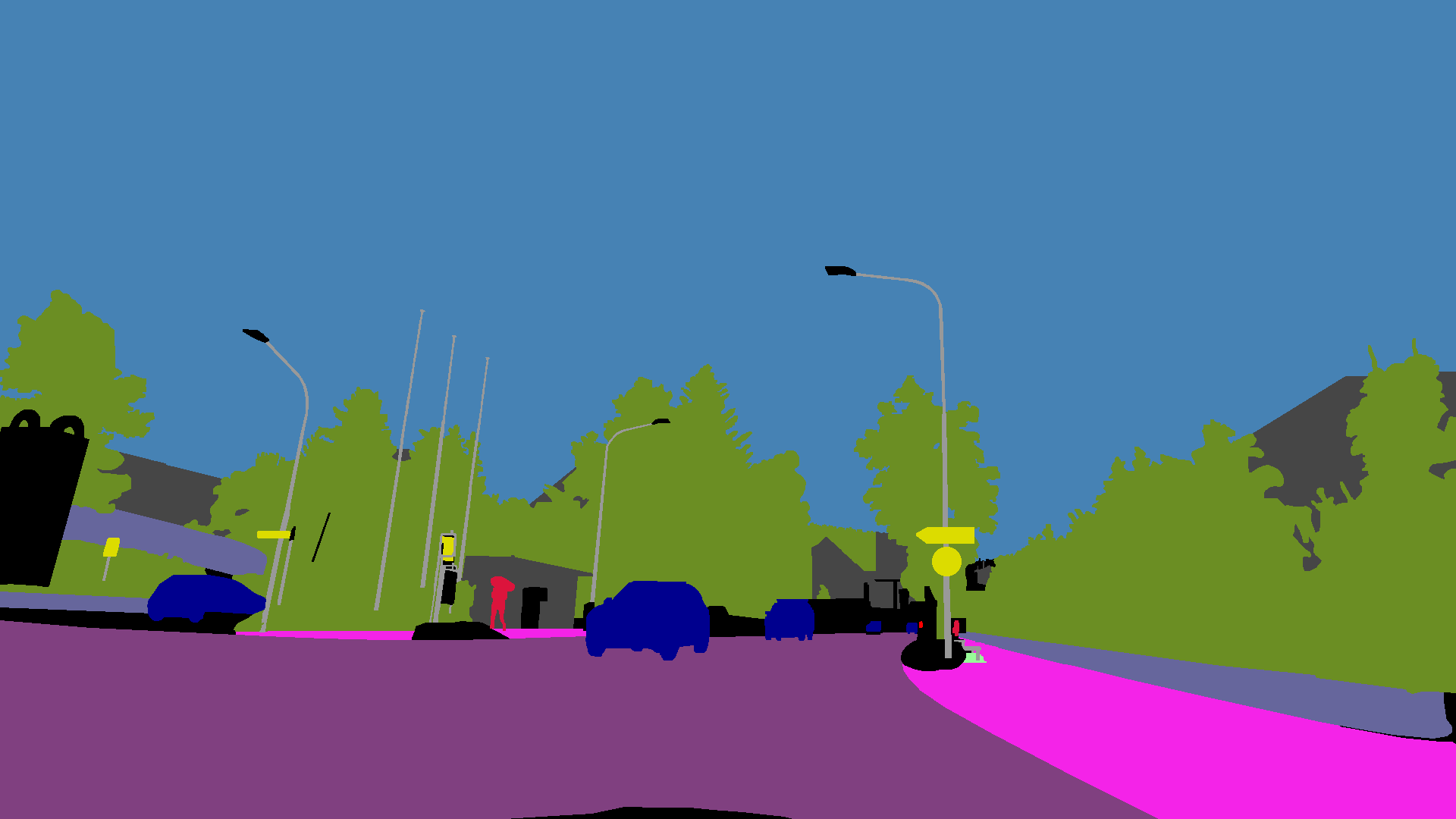} &
        \includegraphics[width=0.23\textwidth]{  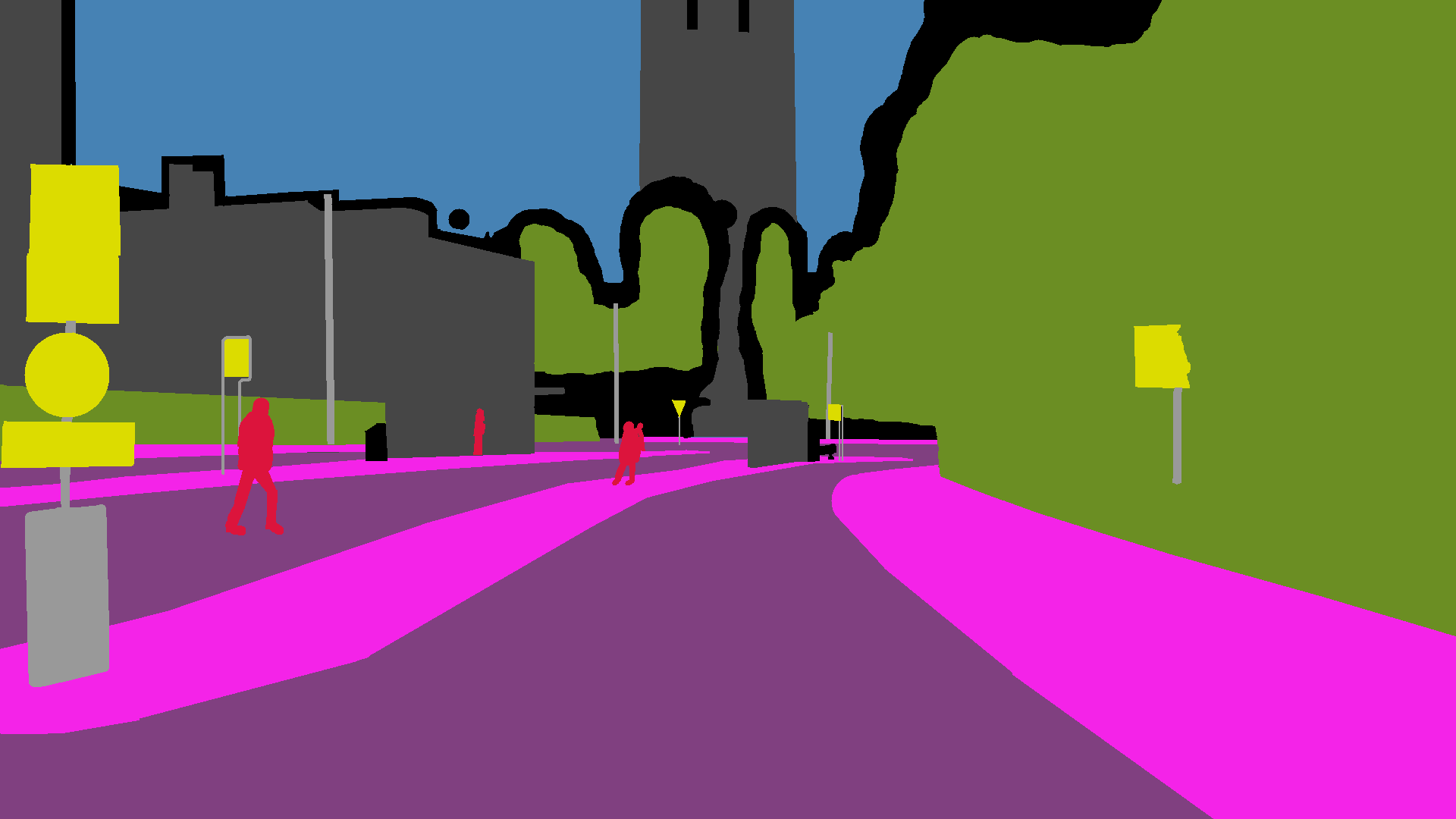} &
        \includegraphics[width=0.23\textwidth]{  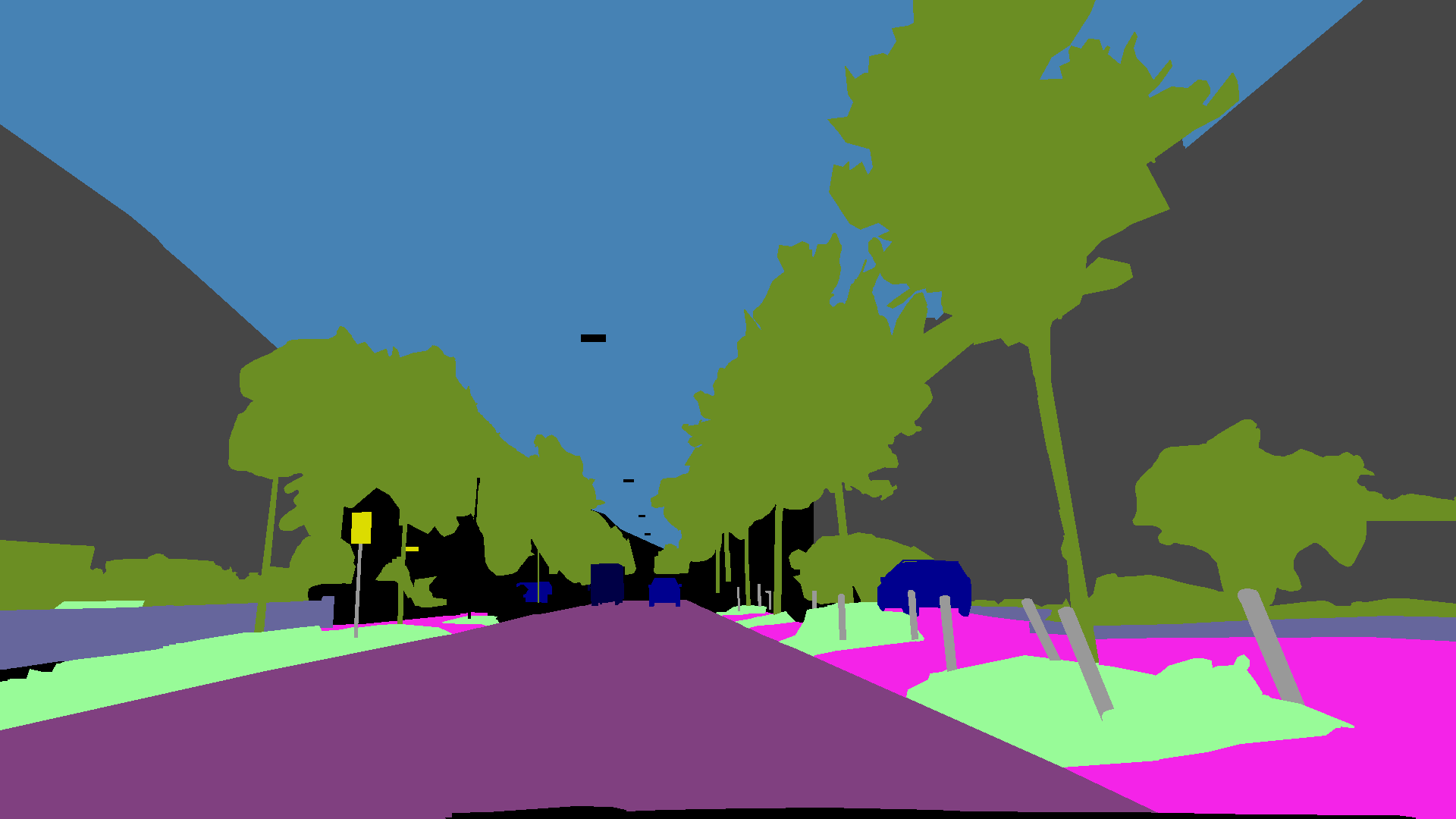} \\
        
        \raisebox{0.0cm}{\begin{sideways}\tiny SegFormer~\cite{xie2021}\end{sideways}} &
        \includegraphics[width=0.23\textwidth]{  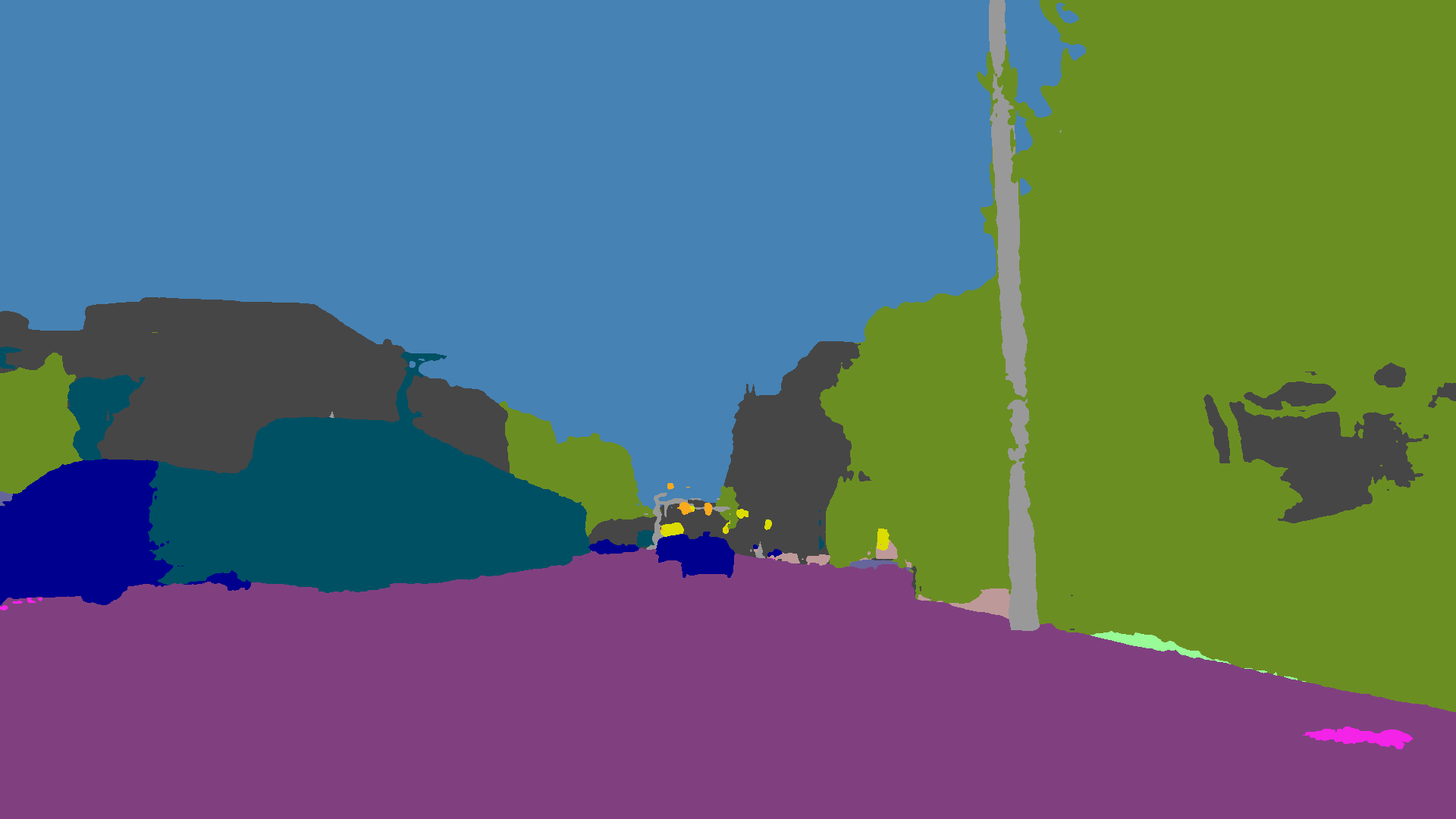} &
        \includegraphics[width=0.23\textwidth]{  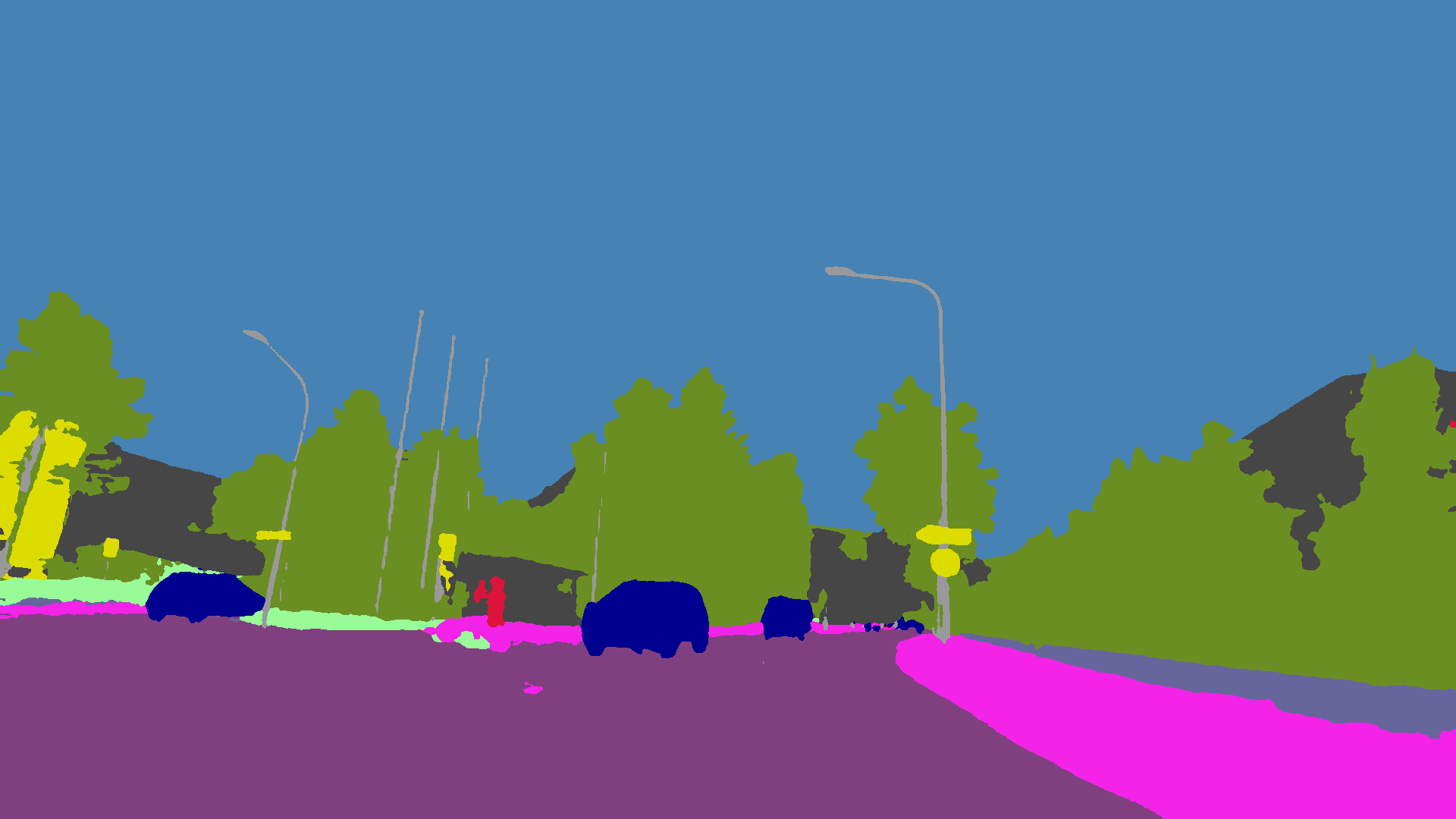} &
        \includegraphics[width=0.23\textwidth]{  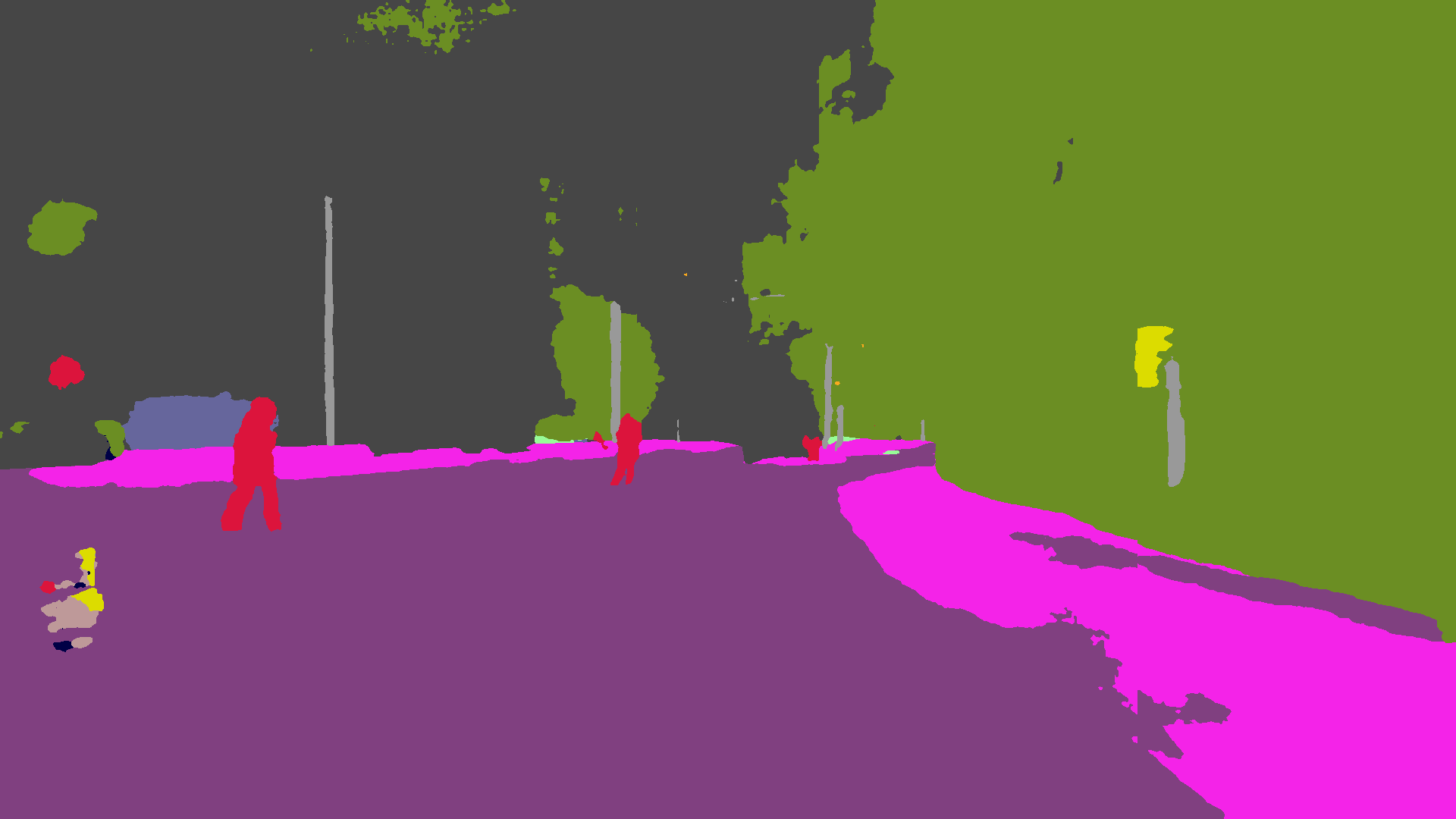} &
        \includegraphics[width=0.23\textwidth]{  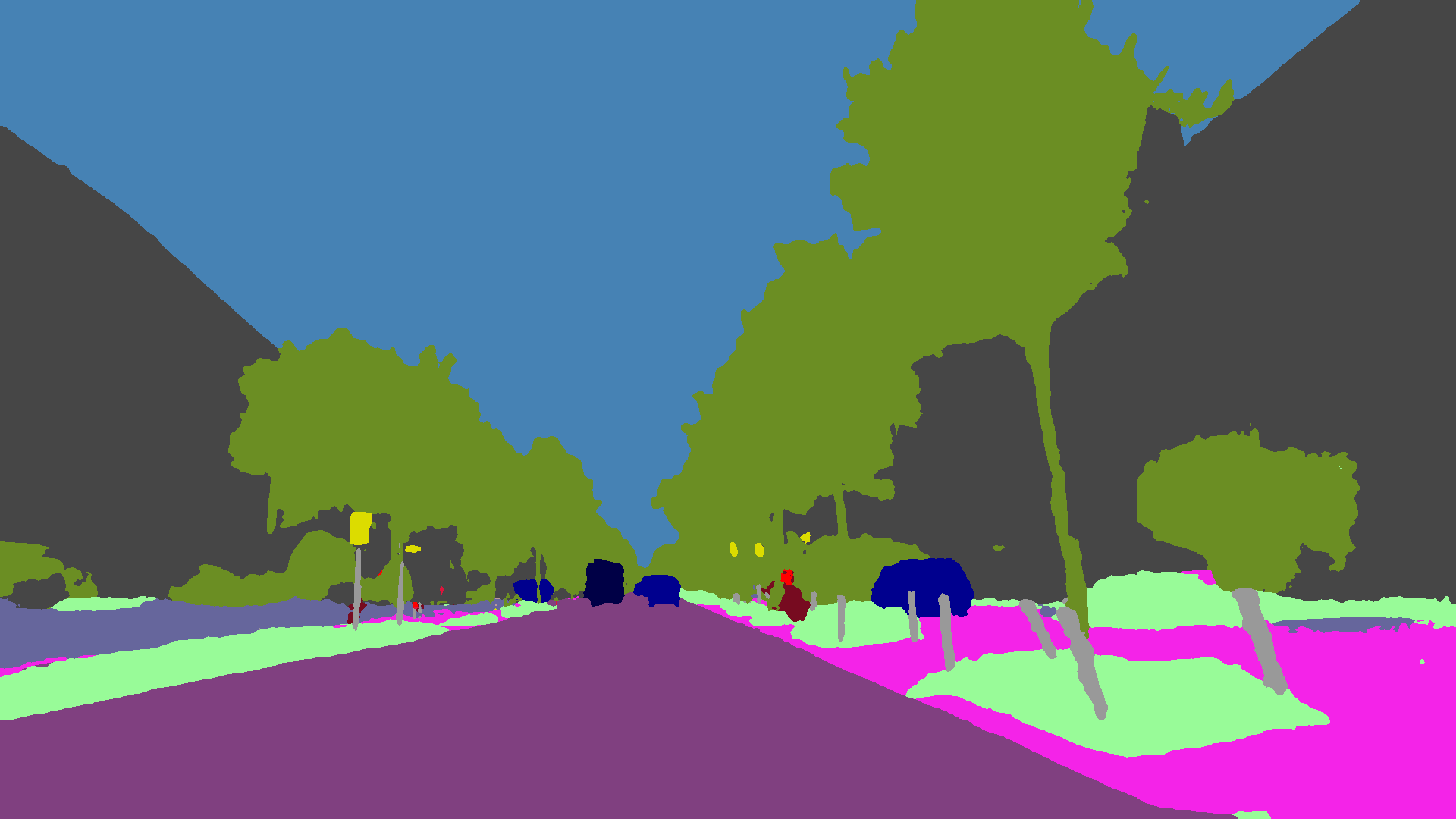} \\

        \raisebox{0.3cm}{\begin{sideways}\tiny Refign~\cite{Bruggemann2023}\end{sideways}} &
        \includegraphics[width=0.23\textwidth]{  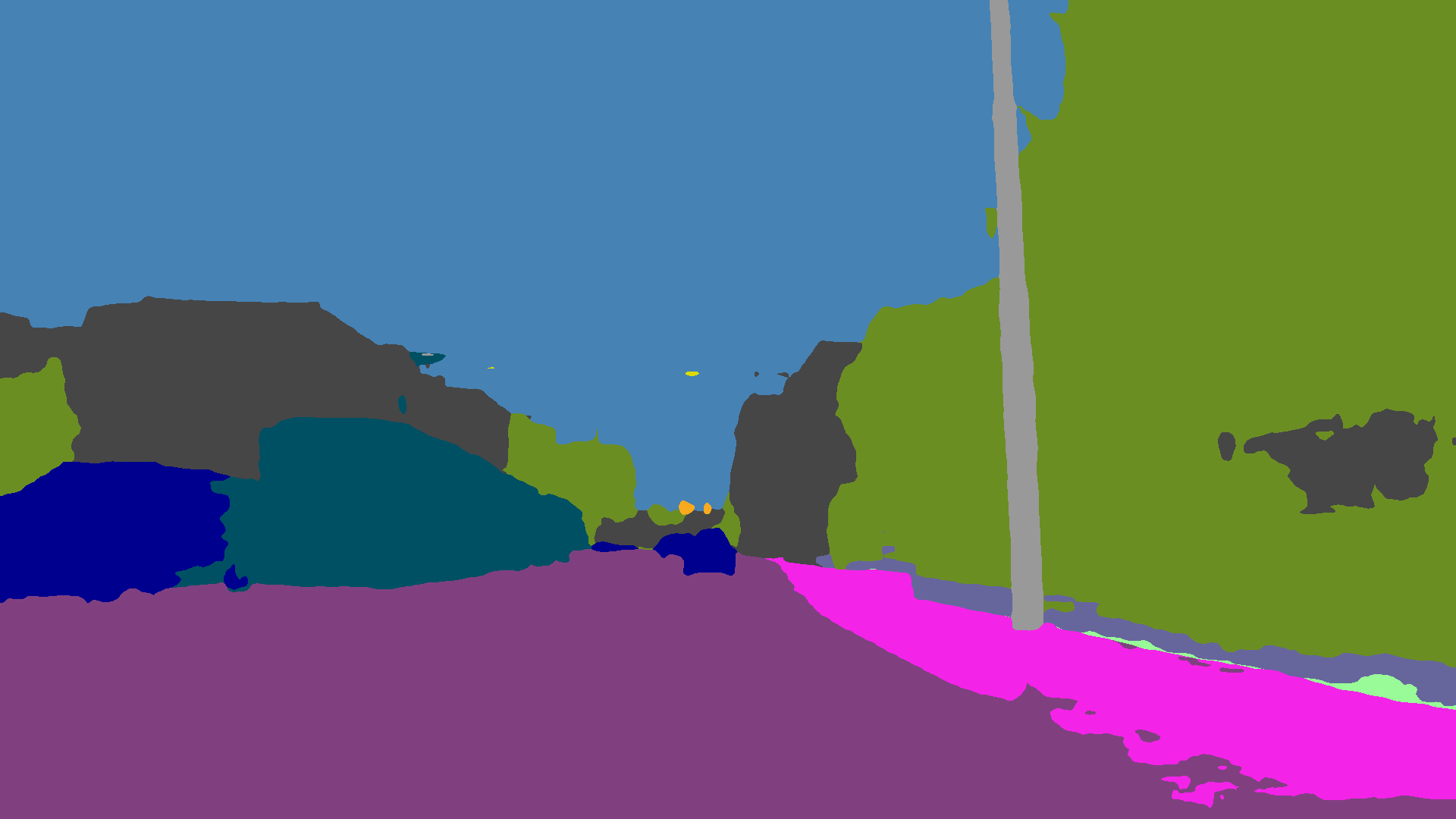} &
        \includegraphics[width=0.23\textwidth]{  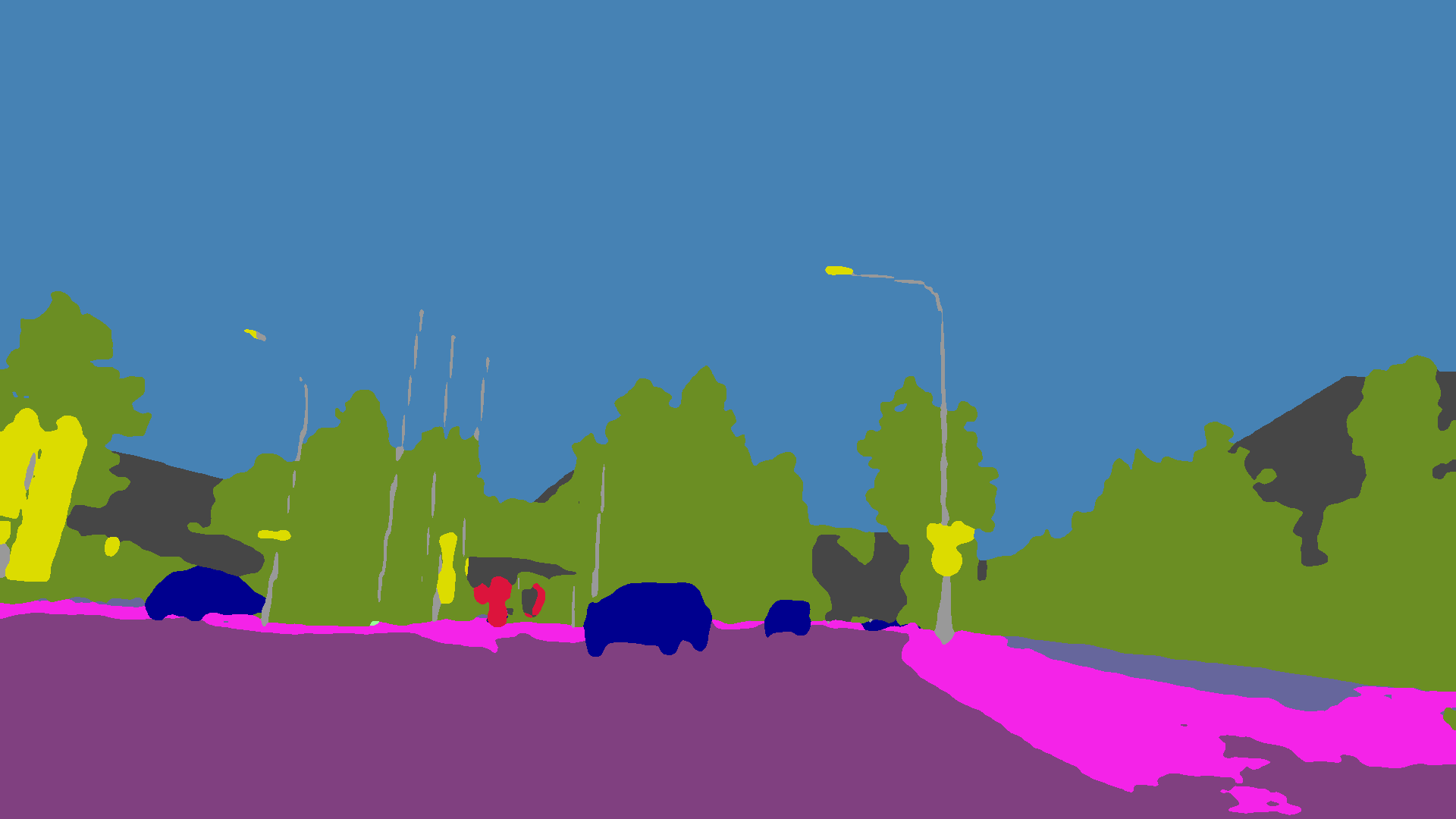} &
        \includegraphics[width=0.23\textwidth]{  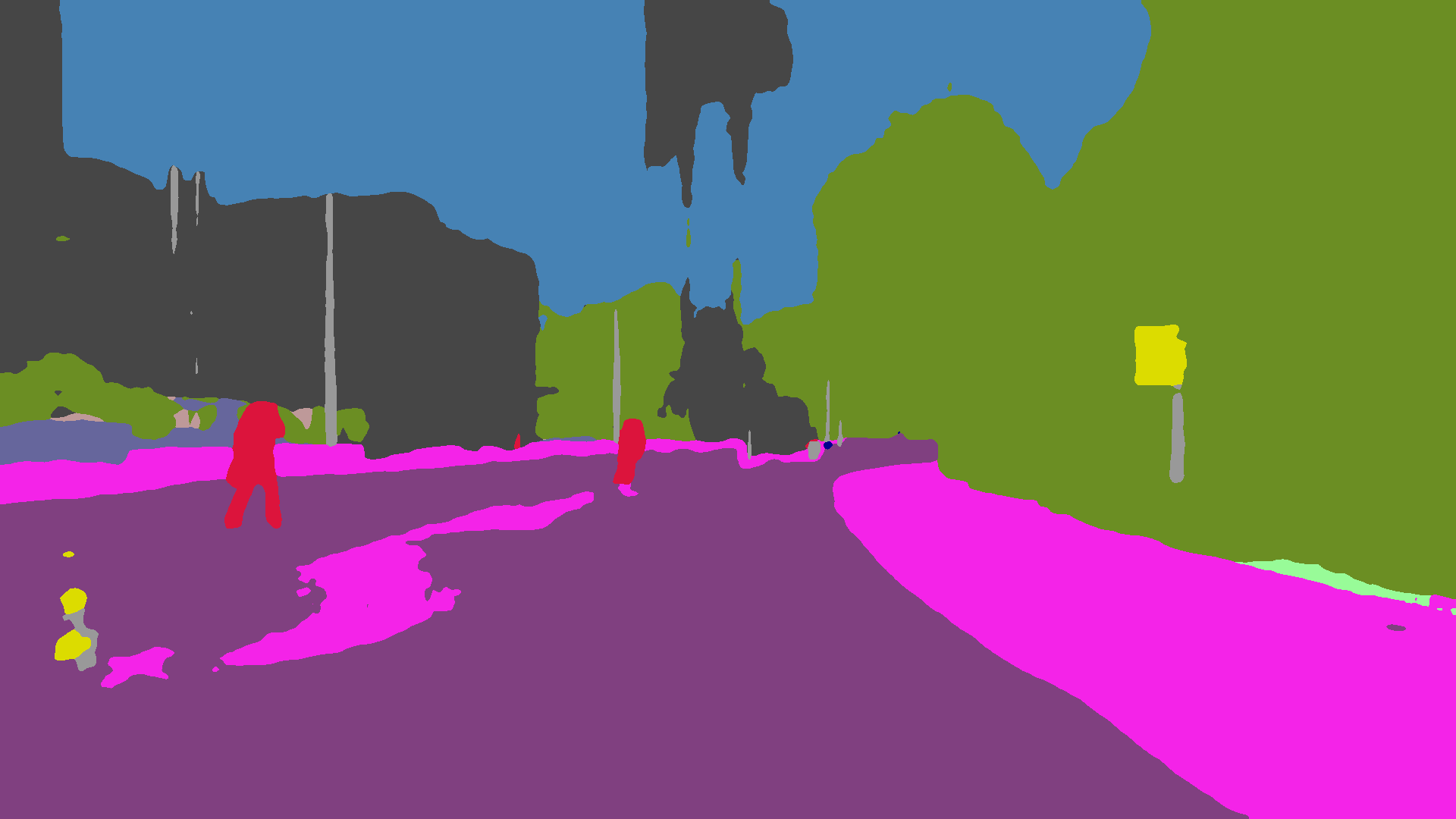} &
        \includegraphics[width=0.23\textwidth]{  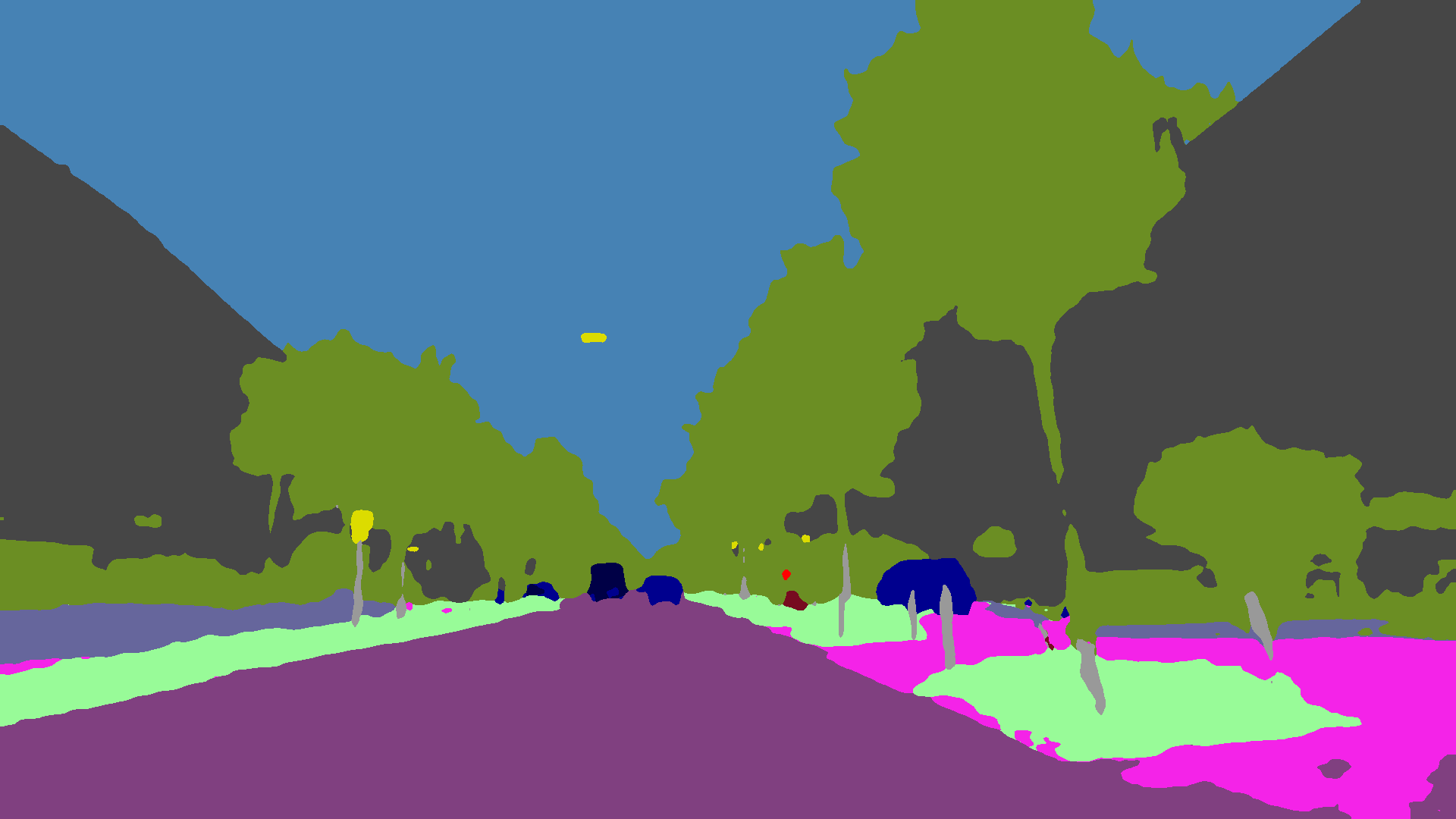} \\

        \raisebox{0.35cm}{\begin{sideways}\tiny CMA~\cite{Bruggemann2023CMA}\end{sideways}} &
        \includegraphics[width=0.23\textwidth]{  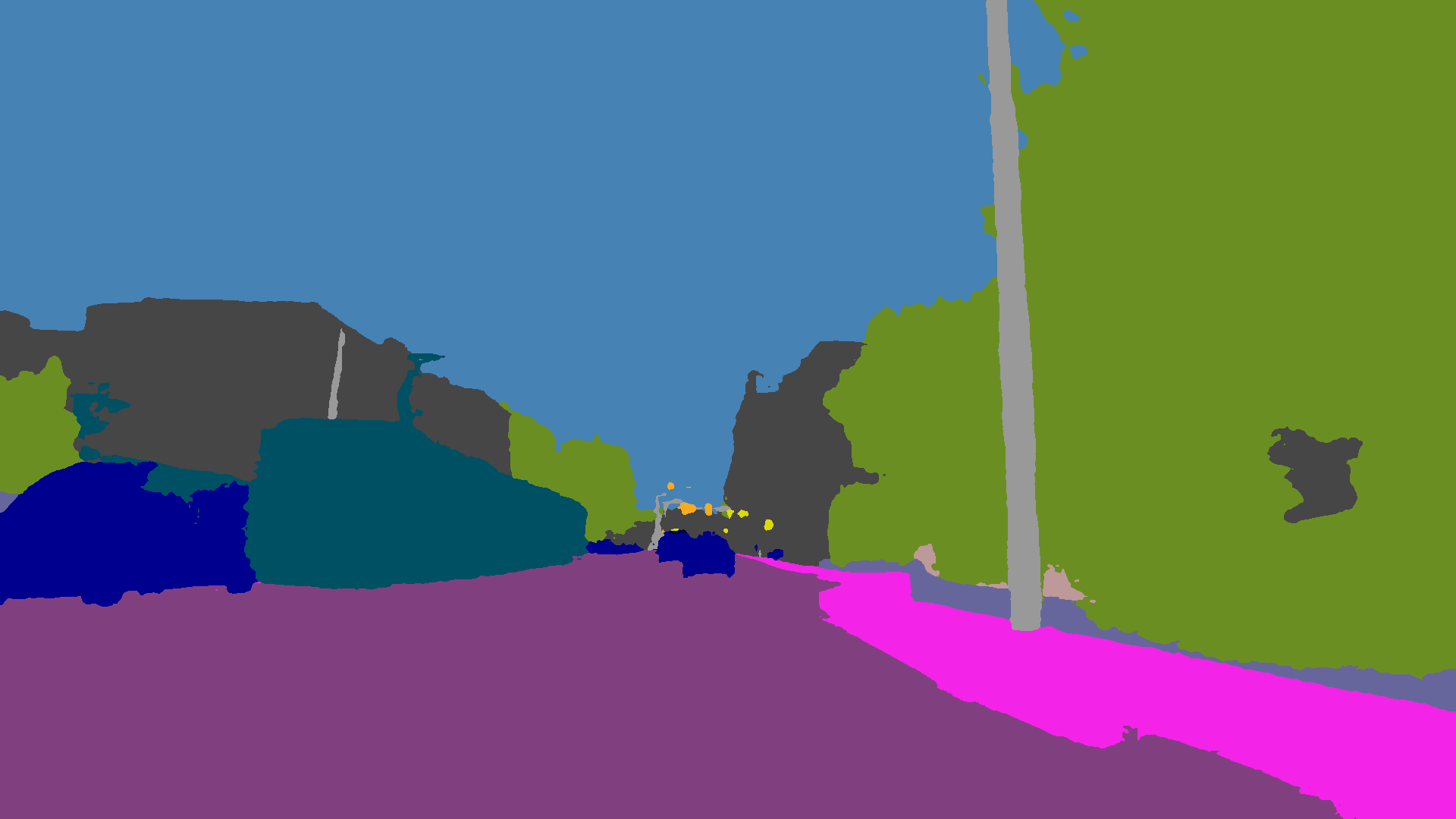} &
        \includegraphics[width=0.23\textwidth]{  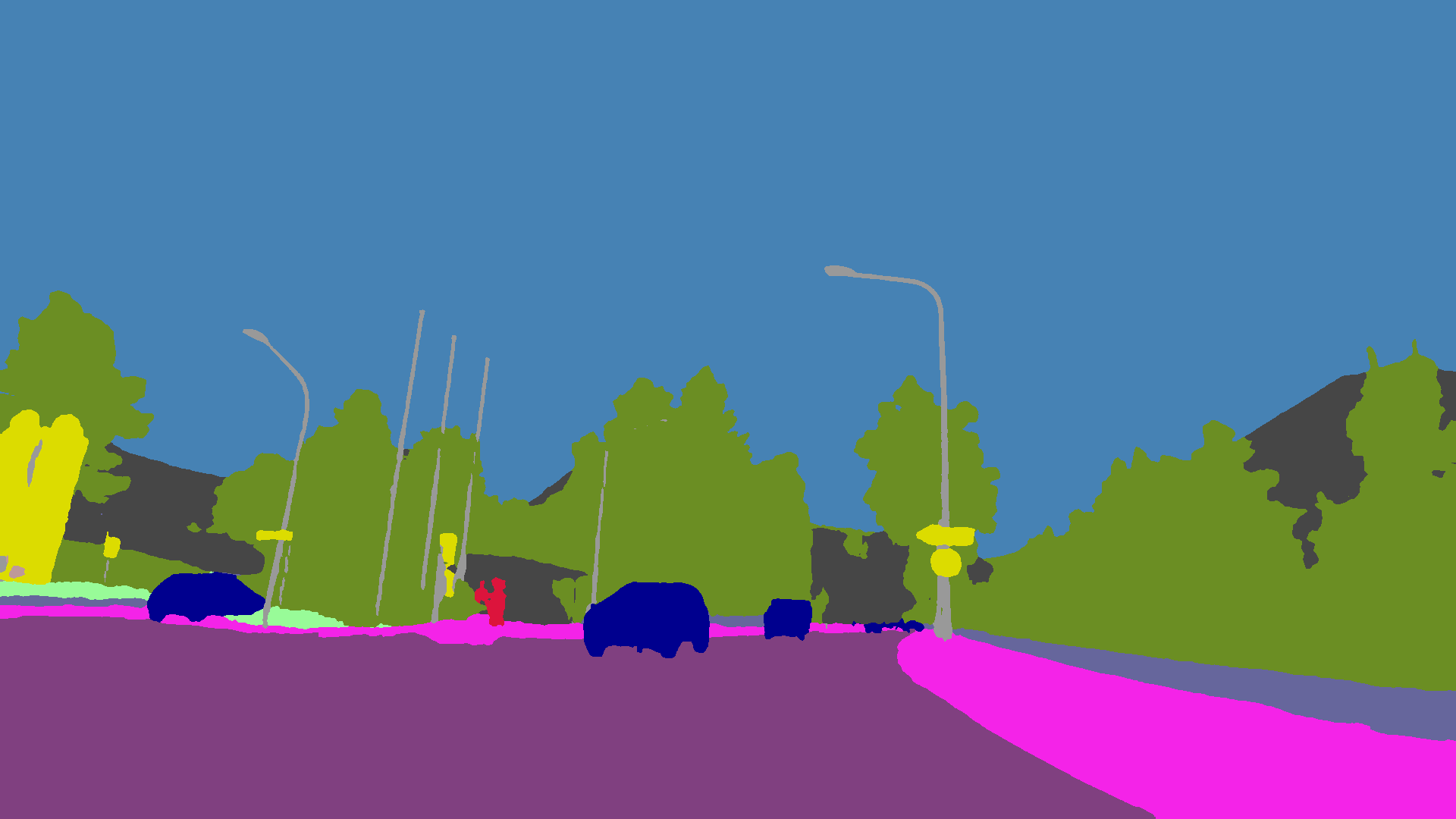} &
        \includegraphics[width=0.23\textwidth]{  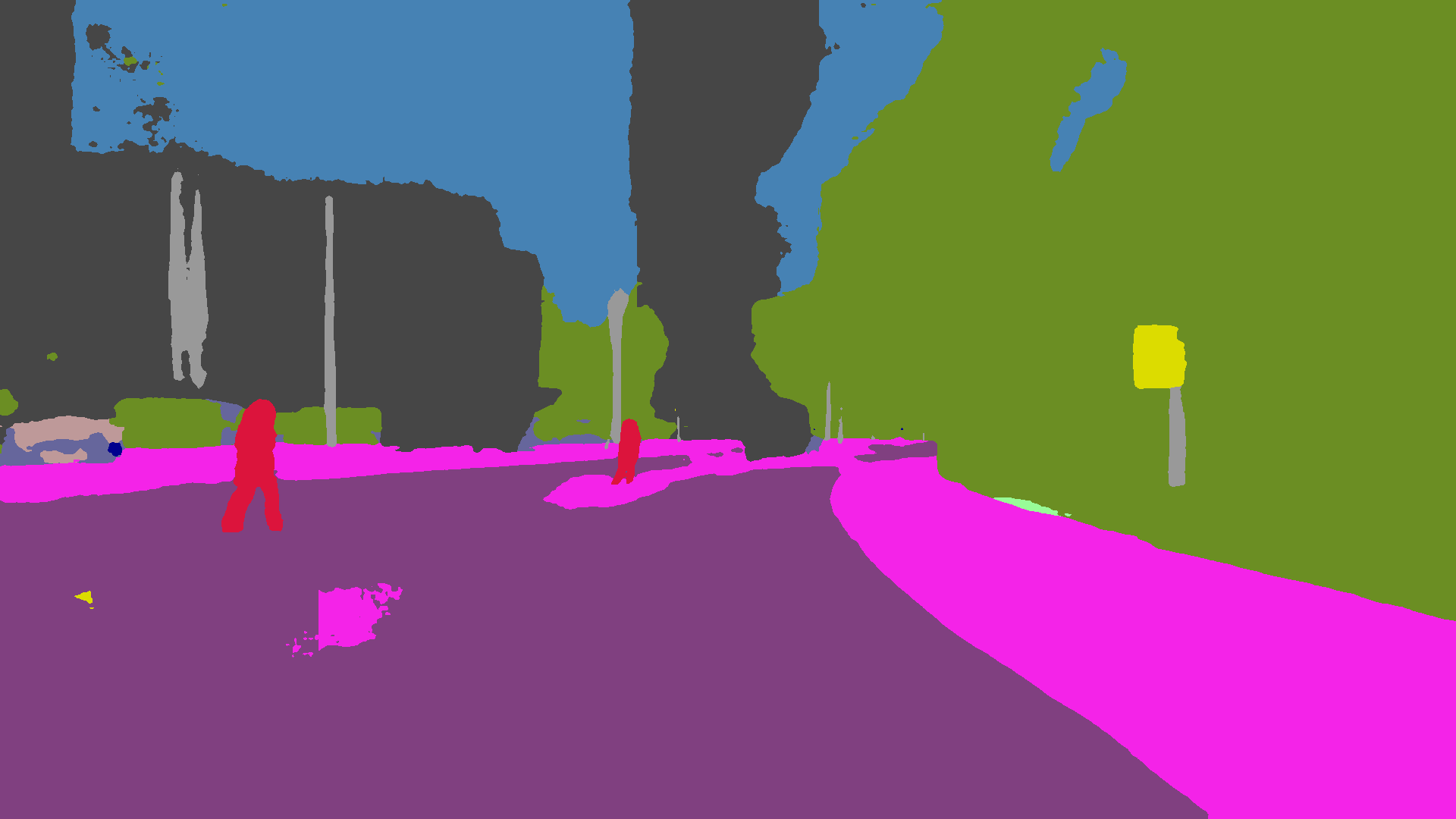} &
        \includegraphics[width=0.23\textwidth]{  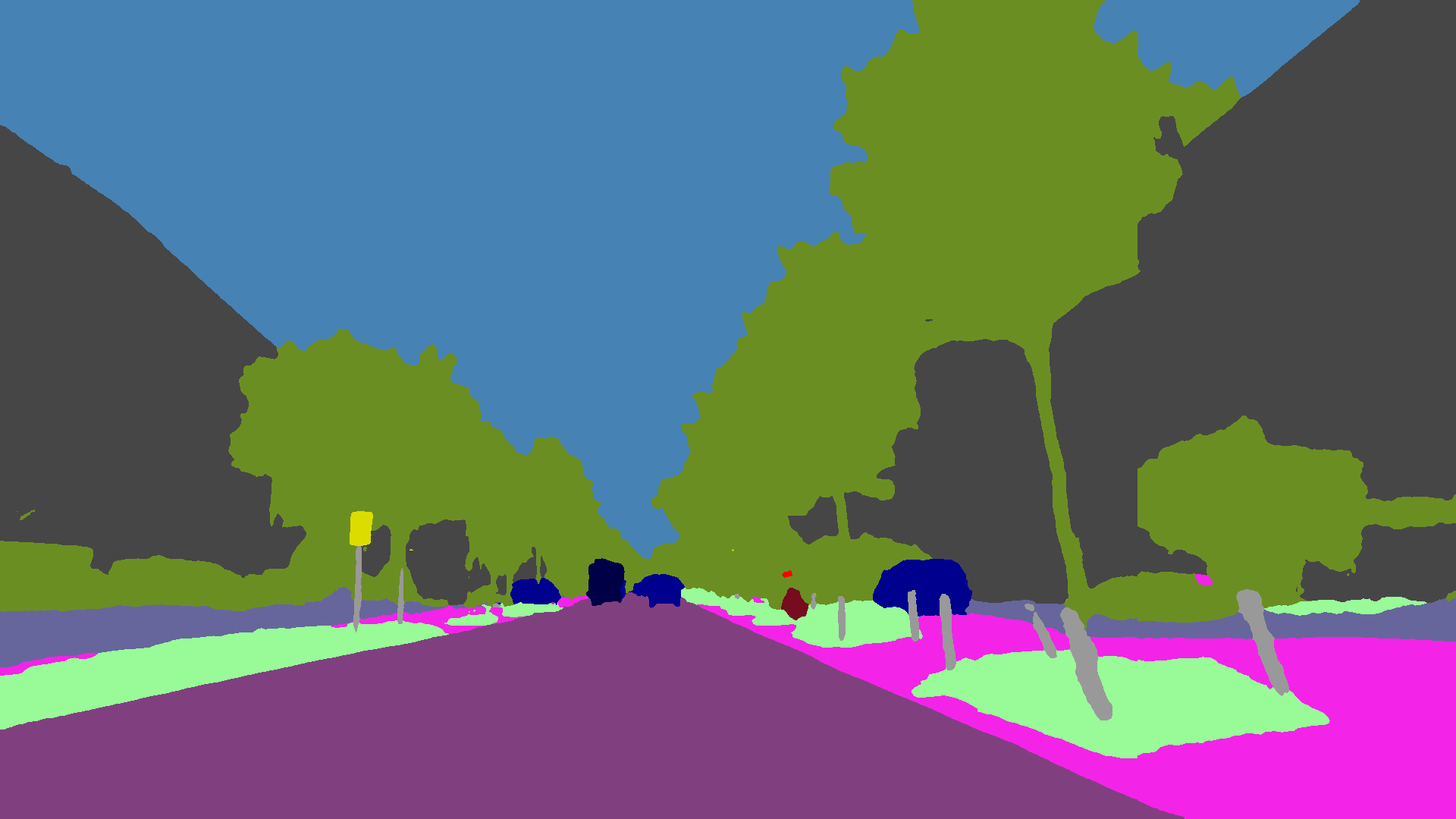} \\

        \raisebox{0.05cm}{\begin{sideways}\tiny BTSeg (Ours)\end{sideways}} &
        \includegraphics[width=0.23\textwidth]{  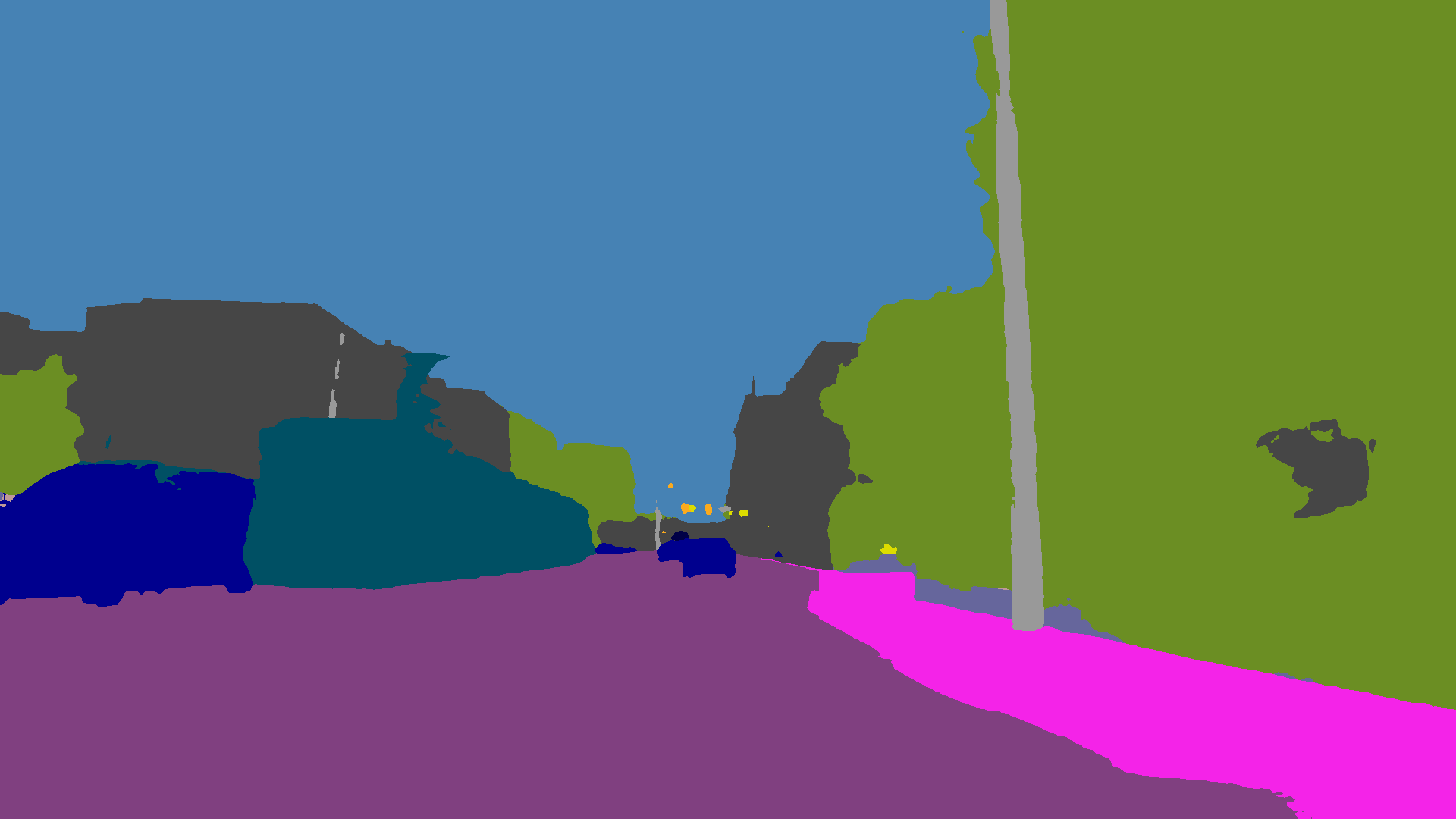} &
        \includegraphics[width=0.23\textwidth]{  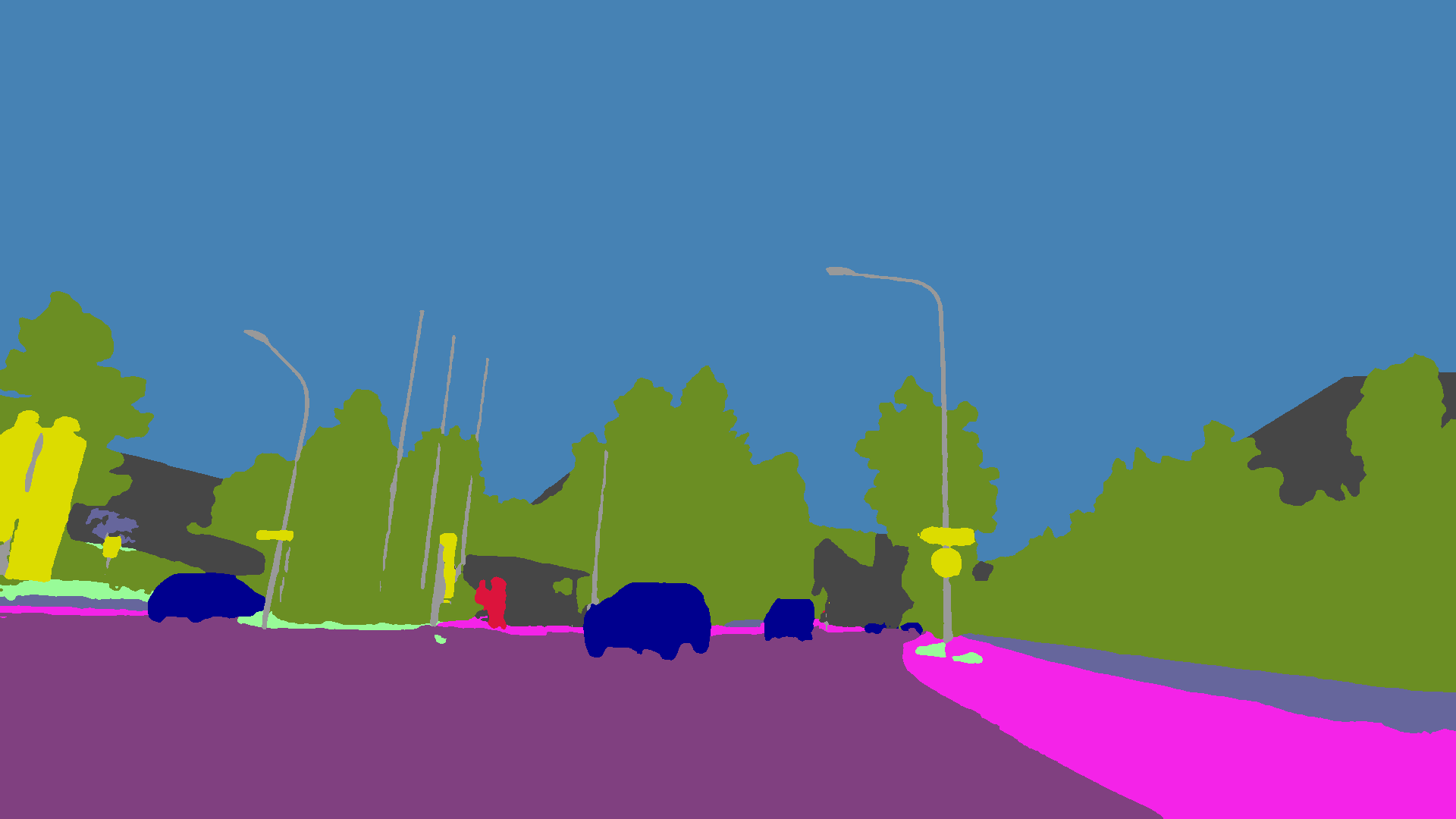} &
        \includegraphics[width=0.23\textwidth]{  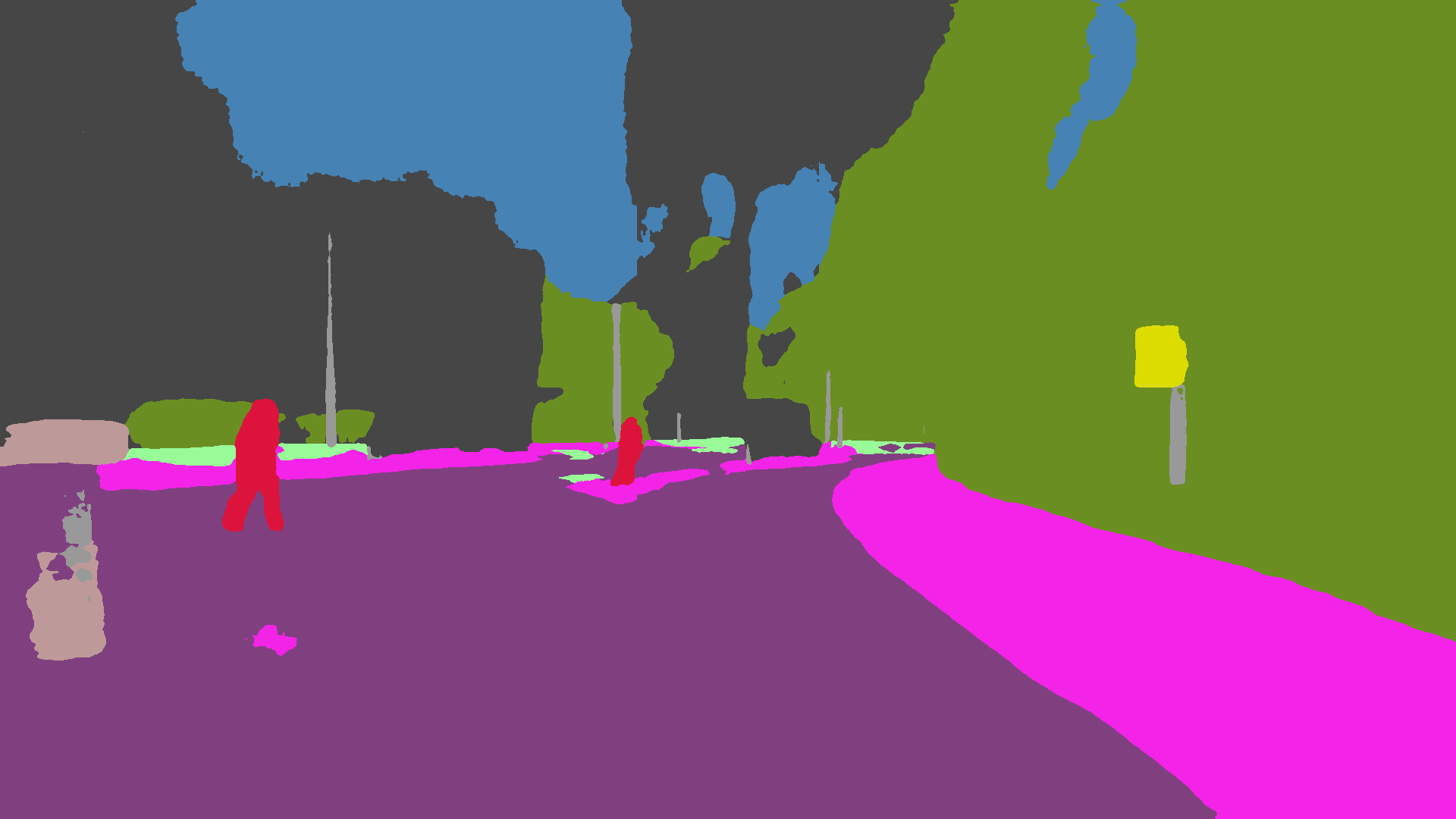} &
        \includegraphics[width=0.23\textwidth]{  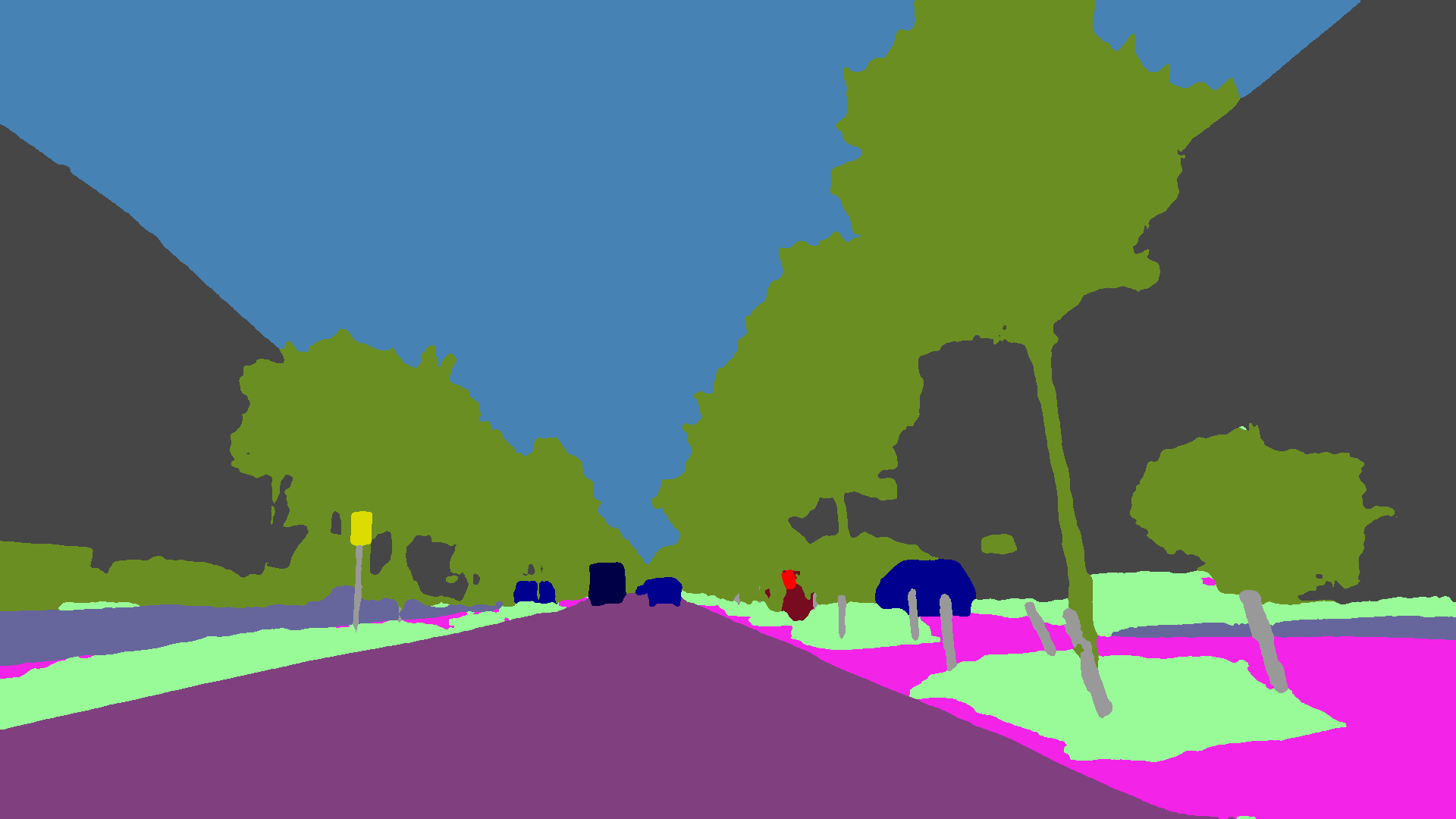} \\
    \end{tabular}

\caption{Qualitative results for randomly selected images from the validation subset of the ACDC dataset~\cite{Sakaridis2021}, showing from left to right snowy, rainy, nighttime, and foggy conditions. Best viewed digitally and zoomed in.}
\label{fig:overview_sota}

\end{figure}

\cref{fig:overview_sota} shows a qualitative comparison of randomly selected images from the ACDC validation subset, aligning with the quantitative evaluation. The improvements from domain adaptation methods are evident, particularly in the segmentation of sidewalks. The differences between Refign, CMA, and BTSeg are subtle, with CMA and BTSeg producing sharper object boundaries than Refign.

\subsection{Ablation Study}
\label{subsec:ablation}

In this section, we analyze and evaluate our design choices.
All numbers are from the ACDC validation set (adverse weather only) and we use the SegFormer configuration described in \cref{subsec:implementationdetails} as foundation, but with a randomized cropping to 512 by 512 and without the resizing to 1080 by 1080, facilitating a more efficient training.

\begin{table}
\centering
\caption{Results of the ablation study performed on the ACDC validation subset reporting the IoU.}
\label{tab:ablation}
\ra{1.3}
%
%
\pgfplotsset{%
    colormap/RdBu-5,
}%
\pgfplotstableset{%
    header=false,
    col sep=&,
    row sep=\\,
    every head row/.style={output empty row},
    every last row/.style={after row=\bottomrule},
    /color cells/min/.initial=0,
    /color cells/max/.initial=1000,
    /color cells/textcolor/.initial=,
    %
    color cells/.code={%
        \pgfqkeys{/color cells}{#1}%
        \pgfkeysalso{%
            postproc cell content/.code={%
                \begingroup
                %
                \pgfkeysgetvalue{/pgfplots/table/@preprocessed cell content}\value
                \ifx\value\empty
                    \endgroup
                \else
                \pgfmathfloatparsenumber{\value}%
                \pgfmathfloattofixed{\pgfmathresult}%
                \let\value=\pgfmathresult
                %
                \pgfplotscolormapaccess
                    [\pgfkeysvalueof{/color cells/min}:\pgfkeysvalueof{/color cells/max}]
                    {\value}
                    {\pgfkeysvalueof{/pgfplots/colormap name}}%
                %
                \pgfkeysgetvalue{/pgfplots/table/@cell content}\typesetvalue
                \pgfkeysgetvalue{/color cells/textcolor}\textcolorvalue
                %
                \toks0=\expandafter{\typesetvalue}%
                \xdef\temp{%
                    \noexpand\pgfkeysalso{%
                        @cell content={%
                            \noexpand\cellcolor[rgb]{\pgfmathresult}%
                            \noexpand\definecolor{mapped color}{rgb}{\pgfmathresult}%
                            \ifx\textcolorvalue\empty
                            \else
                                \noexpand\color{\textcolorvalue}%
                            \fi
                            \the\toks0 %
                        }%
                    }%
                }%
                \endgroup
                \temp
                \fi
            }%
        }%
    }
}%
\def\mycr{20.0}
\pgfmathsetmacro{\roadmin}{86.67 - \mycr}%
\pgfmathsetmacro{\roadmax}{86.67 + \mycr}%
\pgfmathsetmacro{\sidewmin}{51.88 - \mycr}%
\pgfmathsetmacro{\sidewmax}{51.88 + \mycr}%
\pgfmathsetmacro{\buildmin}{73.01 - \mycr}%
\pgfmathsetmacro{\buildmax}{73.01 + \mycr}%
\pgfmathsetmacro{\wallmin}{41.1 - \mycr}%
\pgfmathsetmacro{\wallmax}{41.1 + \mycr}%
\pgfmathsetmacro{\fencemin}{36.76 - \mycr}%
\pgfmathsetmacro{\fencemax}{36.76 + \mycr}%
\pgfmathsetmacro{\polemin}{44.41 - \mycr}%
\pgfmathsetmacro{\polemax}{44.41 + \mycr}%
\pgfmathsetmacro{\lightmin}{60.4 - \mycr}%
\pgfmathsetmacro{\lightmax}{60.4 + \mycr}%
\pgfmathsetmacro{\signmin}{57.08 - \mycr}%
\pgfmathsetmacro{\signmax}{57.08 + \mycr}%
\pgfmathsetmacro{\vegetmin}{75.73 - \mycr}%
\pgfmathsetmacro{\vegetmax}{75.73 + \mycr}%
\pgfmathsetmacro{\terrainmin}{29.91 - \mycr}%
\pgfmathsetmacro{\terrainmax}{29.91 + \mycr}%
\pgfmathsetmacro{\skymin}{96.79 - \mycr}%
\pgfmathsetmacro{\skymax}{96.79 + \mycr}%
\pgfmathsetmacro{\personmin}{57.4 - \mycr}%
\pgfmathsetmacro{\personmax}{57.4 + \mycr}%
\pgfmathsetmacro{\ridermin}{44.86 - \mycr}%
\pgfmathsetmacro{\ridermax}{44.86 + \mycr}%
\pgfmathsetmacro{\carmin}{74.66 - \mycr}%
\pgfmathsetmacro{\carmax}{74.66 + \mycr}%
\pgfmathsetmacro{\truckmin}{66.62 - \mycr}%
\pgfmathsetmacro{\truckmax}{66.62 + \mycr}%
\pgfmathsetmacro{\busmin}{59.42 - \mycr}%
\pgfmathsetmacro{\busmax}{59.42 + \mycr}%
\pgfmathsetmacro{\trainmin}{59.76 - \mycr}%
\pgfmathsetmacro{\trainmax}{59.76 + \mycr}%
\pgfmathsetmacro{\motorcmin}{45.37 - \mycr}%
\pgfmathsetmacro{\motorcmax}{45.37 + \mycr}%
\pgfmathsetmacro{\bicyclemin}{45.23 - \mycr}%
\pgfmathsetmacro{\bicyclemax}{45.23 + \mycr}%
\pgfmathsetmacro{\meanmin}{58.32 - \mycr}%
\pgfmathsetmacro{\meanmax}{58.32 + \mycr}%
\pgfplotstableread[header=true]{
mean & road & sidew. & build. & wall & fence & pole & light & sign & veget. & terrain & sky & person & rider & car & truck & bus & train & motorc. & bicycle \\
58.32 & 86.67 & 51.88 & 73.01 & 41.1 & 36.76 & 44.41 & 60.4 & 57.08 & 75.73 & 29.91 & 96.79 & 57.4 & 44.86 & 74.66 & 66.62 & 59.42 & 59.76 & 45.37 & 45.23\\     
64.18 & 90.33 & 67.43 & 82.94 & 47.50 & 39.14 & 56.07 & 74.11 & 59.40 & 78.08 & 39.77 & 87.63 & 62.97 & 37.88 & 84.32 & 70.23 & 79.51 & 80.73 & 46.06 & 35.26 \\ 
65.46 & 91.08 & 69.50 & 83.47 & 49.31 & 37.53 & 58.67 & 73.90 & 60.31 & 79.95 & 42.51 & 88.76 & 61.25 & 42.50 & 85.22 & 70.93 & 86.78 & 82.63 & 48.23 & 31.26 \\ 
65.71 & 90.11 & 66.50 & 84.25 & 48.49 & 39.67 & 58.16 & 72.48 & 61.11 & 80.16 & 42.86 & 89.18 & 59.48 & 40.28 & 85.39 & 72.57 & 86.18 & 84.43 & 49.34 & 37.80 \\ 
65.42 & 89.74 & 67.31 & 83.85 & 48.88 & 38.07 & 59.33 & 73.19 & 60.74 & 80.23 & 42.83 & 88.51 & 59.65 & 40.34 & 85.13 & 75.41 & 86.07 & 81.32 & 48.84 & 33.50 \\ 
65.23 & 89.74 & 66.01 & 84.02 & 49.62 & 39.34 & 58.70 & 74.01 & 61.39 & 79.78 & 43.65 & 88.62 & 60.66 & 42.58 & 84.69 & 73.46 & 85.46 & 72.26 & 49.43 & 35.91 \\ 
66.20 & 90.07 & 67.80 & 84.17 & 50.50 & 39.00 & 59.92 & 74.03 & 61.40 & 80.32 & 43.21 & 88.63 & 61.04 & 44.23 & 84.86 & 76.50 & 86.96 & 83.53 & 49.02 & 32.54 \\ 
}\mytable%
\def\mycolw{\textnormal{\textsc{al}} }
\resizebox{\textwidth}{!}{
\begin{tabular}{@{}c@{}c@{}c@{}c@{}c@{}c@{}c@{}c@{}c@{}c@{}c@{}c@{}c@{}c@{}c@{}c@{}c@{}c@{}c@{}c@{}c@{}c@{}c@{}c@{}c@{}r@{}}\toprule
\multicolumn{4}{c}{\textbf{Methods}} & \multicolumn{20}{c}{\textbf{IoU}$\uparrow$} \\
\cmidrule{5-24}
\rotside{$\mathcal{L}_\text{BT}$} & \rotside{warp} & \rotside{crop} & \rotside{pooling} & \rotatebox[origin=c]{90}{\textbf{mean}} & \rotatebox[origin=c]{90}{road} & \rotatebox[origin=c]{90}{sidew.} & \rotatebox[origin=c]{90}{build.} & \rotatebox[origin=c]{90}{wall} & \rotatebox[origin=c]{90}{fence} & \rotatebox[origin=c]{90}{pole} & \rotatebox[origin=c]{90}{light} & \rotatebox[origin=c]{90}{sign} & \rotatebox[origin=c]{90}{veget.} & \rotatebox[origin=c]{90}{terrain} & \rotatebox[origin=c]{90}{sky} & \rotatebox[origin=c]{90}{person} & \rotatebox[origin=c]{90}{rider} & \rotatebox[origin=c]{90}{car} & \rotatebox[origin=c]{90}{truck} & \rotatebox[origin=c]{90}{bus} & \rotatebox[origin=c]{90}{train} & \rotatebox[origin=c]{90}{motorc.} & \rotatebox[origin=c]{90}{bicycle} \\ \bottomrule
{\pgfplotstabletypeset[string type]{
\textover[c]{\phantom{\checkmark}}{\mycolw} \\ 
\checkmark \\ 
\checkmark \\
\checkmark \\
\checkmark \\
\checkmark \\
\checkmark \\
}} &
{\pgfplotstabletypeset[string type]{
\textover[c]{\phantom{\checkmark}}{\mycolw} \\ 
\phantom{\checkmark} \\ 
\checkmark \\
\checkmark \\
\checkmark \\
\checkmark \\
\checkmark \\
}} &
{\pgfplotstabletypeset[string type]{
\textover[c]{\phantom{\checkmark}}{\mycolw} \\ 
\phantom{\checkmark} \\ 
\phantom{\checkmark} \\ 
\checkmark \\
\checkmark \\
\checkmark \\
\checkmark \\
}} &
{\pgfplotstabletypeset[string type]{
\textover[c]{--}{\mycolw} \\ 
Avg. \\ 
Avg. \\ 
Avg. \\ 
Segm. \\ 
Conf. \\
SegConf. \\
}} &
{\pgfplotstabletypeset[
    columns={mean},
    color cells={min=\meanmin,max=\meanmax},
    fixed zerofill,
    precision=1,
    columns/0/.style={column type=r},%
]\mytable} &
{\pgfplotstabletypeset[
    columns={road},
    color cells={min=\roadmin,max=\roadmax},
    fixed zerofill,
    precision=1,
]\mytable} &
{\pgfplotstabletypeset[
    columns={sidew.},
    color cells={min=\sidewmin,max=\sidewmax},
    fixed zerofill,
    precision=1,
]\mytable} &
{\pgfplotstabletypeset[
    columns={build.},
    color cells={min=\buildmin,max=\buildmax},
    fixed zerofill,
    precision=1,
]\mytable} &
{\pgfplotstabletypeset[
    columns={wall},
    color cells={min=\wallmin,max=\wallmax},
    fixed zerofill,
    precision=1,
]\mytable} &
{\pgfplotstabletypeset[
    columns={fence},
    color cells={min=\fencemin,max=\fencemax},
    fixed zerofill,
    precision=1,
]\mytable} &
{\pgfplotstabletypeset[
    columns={pole},
    color cells={min=\polemin,max=\polemax},
    fixed zerofill,
    precision=1,
]\mytable} &
{\pgfplotstabletypeset[
    columns={light},
    color cells={min=\lightmin,max=\lightmax},
    fixed zerofill,
    precision=1,
]\mytable} &
{\pgfplotstabletypeset[
    columns={sign},
    color cells={min=\signmin,max=\signmax},
    fixed zerofill,
    precision=1,
]\mytable} &
{\pgfplotstabletypeset[
    columns={veget.},
    color cells={min=\vegetmin,max=\vegetmax},
    fixed zerofill,
    precision=1,
]\mytable} &
{\pgfplotstabletypeset[
    columns={terrain},
    color cells={min=\terrainmin,max=\terrainmax},
    fixed zerofill,
    precision=1,
]\mytable} &
{\pgfplotstabletypeset[
    columns={sky},
    color cells={min=\skymin,max=\skymax},
    fixed zerofill,
    precision=1,
]\mytable} &
{\pgfplotstabletypeset[
    columns={person},
    color cells={min=\personmin,max=\personmax},
    fixed zerofill,
    precision=1,
]\mytable} &
{\pgfplotstabletypeset[
    columns={rider},
    color cells={min=\ridermin,max=\ridermax},
    fixed zerofill,
    precision=1,
]\mytable} &
{\pgfplotstabletypeset[
    columns={car},
    color cells={min=\carmin,max=\carmax},
    fixed zerofill,
    precision=1,
]\mytable} &
{\pgfplotstabletypeset[
    columns={truck},
    color cells={min=\truckmin,max=\truckmax},
    fixed zerofill,
    precision=1,
]\mytable} &
{\pgfplotstabletypeset[
    columns={bus},
    color cells={min=\busmin,max=\busmax},
    fixed zerofill,
    precision=1,
]\mytable} &
{\pgfplotstabletypeset[
    columns={train},
    color cells={min=\trainmin,max=\trainmax},
    fixed zerofill,
    precision=1,
]\mytable} &
{\pgfplotstabletypeset[
    columns={motorc.},
    color cells={min=\motorcmin,max=\motorcmax},
    fixed zerofill,
    precision=1,
]\mytable} &
{\pgfplotstabletypeset[
    columns={bicycle},
    color cells={min=\bicyclemin,max=\bicyclemax},
    fixed zerofill,
    precision=1,
]\mytable} %
\end{tabular}
}
\end{table}

\cref{tab:ablation} presents the IoU metrics for the different classes present in the ACDC dataset and the overall mean IoU, respectively.
The first row illustrates the results derived from training exclusively with the labeled source images and evaluation on the target images.
This yields an average IoU of \SI{58.32}{\percent} (training SegFormer with the aligned source and target samples and the source segmentation map results in a slightly improved mean IoU of \SI{60.7}{\percent}).
The second row reports the results with Barlow Twins regularization to enforce the encoder to be robust against the appearance changes caused by the different weather conditions, which notably improves the mean IoU by \SI{5.86}{\pp}.
Incorporating the warping / alignment between source and target image improves the overall result to \SI{65.5}{\percent} (third row).
The cropping of invalid image regions post-warping yields an additional slight improvement.
Interestingly, the result slightly decreases when we add segmentation or confidence guided averaging.
For the later, this may be attributed to the conservative confidence estimates in homogeneous image regions, potentially hindering the training.
The segmentation guided averaging is of course susceptible to misclassified image regions, and obstructs the learning of robust encoding, especially if there is coincidental alignment of moving objects.
Finally, the combination of both methods results in the best performing configuration with a mean IoU of \SI{66.2}{\percent}.

In summary, the results show the intrinsic robustness of the proposed Barlow Twins regularization against slight misalignments, which could be interpreted as natural augmentations additional to the appearance changes caused be the weather conditions.

\subsection{Generalization and Robustness}
\label{subsec:acg_benchmark}

Bruggemann~\etal~\cite{Bruggemann2023CMA} introduced the ACG benchmark to evaluate network robustness and generalization after domain adaptation from normal to adverse conditions.
\cref{tab:acg} presents the IoU of recent works after domain adaptation from Cityscapes to ACDC, with SegFormer~\cite{xie2021}, trained solely on Cityscapes, serving as a baseline.
As expected, all domain-adaptation approaches show significant improvement, particularly for the challenging night subset.
Our approach outperforms in all categories, except within the fog subset, showcasing its robustness and strong generalization capabilities.
Notably, it significantly improves over Refign, emphasizing the value of integrating unsupervised learning methods like contrastive learning (CMA~\cite{Bruggemann2023CMA}) and our use of the Barlow Twins loss. 
Despite BTSeg's substantial improvement over Refign and CMA, the night subset remains the most challenging. 
It's essential to note that the night subset introduces a heightened challenge, incorporating images from other domains captured at nighttime (e.g., rain at night), absent in the ACDC dataset.
In conclusion, our method demonstrates robust performance in the presence of unseen data and even more challenging weather conditions.

\begin{table}[t]
\centering
\scriptsize
\caption{Comparison against the state of the art on the ACG benchmark \cite{Bruggemann2023CMA}, for the evaluation of robustness and generalization.}
\label{tab:acg}
\resizebox{\linewidth}{!}{%
\begin{tblr}{
  width = \linewidth,
  colspec = {Q[254]Q[121]Q[131]Q[121]Q[140]Q[121]},
  column{even} = {c},
  column{3} = {c},
  column{5} = {c},
  cell{1}{2} = {c=5}{0.634\linewidth},
  hline{1,7} = {-}{0.08em},
  hline{2} = {2-6}{},
  hline{3,6} = {-}{},
}
\SetCell[r=2]{c}{\textbf{Method}} & \textbf{IoU} $ \uparrow $ &  &  &  & \\
 & fog & night & rain & snow & all\\
SegFormer~\cite{xie2021} & 54.1 & 28.0 & 47.6 & 41.3 & 40.2\\
Refign~\cite{Bruggemann2023} & 54.2 & 37.8 & 56.2 & 49.3 & 48.1\\
CMA~\cite{Bruggemann2023CMA} & 59.7 & 40.0 & 59.6 & \textbf{52.2} & 51.3\\
BTSeg (Ours) & \textbf{61.4} & \textbf{40.2} & \textbf{60.2} & 50.4 & \textbf{51.8}
\end{tblr}
}
\end{table}

\paragraph{Limitations} While our approach proves effective, it does have certain limitations. The necessity for a large batch size in loss computation has been addressed through the implementation of an embedding caching mechanism.
Simultaneously, the loss formulation removes the need for negative samples, distinguishing it from contrastive methods like CMA~\cite{Bruggemann2023CMA}.
Although the training time extends to 48 hours on a single GPU, this is inversely proportional to the number of GPUs employed. Furthermore, the requirement for image pairs during training, facilitating the network in learning a shared latent representation, presents a constraint.
However, it is worth noting that acquiring such pairs is often more practical than obtaining labeled data for the target domain.

\section{Conclusion}
\label{sec:conclusion}

In this paper, we introduced BTSeg, a weakly-supervised domain adaptation method designed for semantically segmenting images captured under challenging weather conditions. Our approach treats images of a specific location in different weather conditions as variations of an underlying, shared base image. Through the incorporation of the Barlow Twins loss, we compel the segmentation network to withstand these assumed augmentations, effectively adapting to diverse weather conditions. Our method leverages pairs of adverse and clear weather images, requiring a pixel-segmentation mask only for the latter, offering a significant advantage in terms of data collection efficiency. Moreover, the utilization of the Barlow Twins loss eliminates the need for selecting negative samples during training.

This paper is the first to demonstrate that Barlow Twins, with specific design choices, can achieve state-of-the-art results in weakly-supervised domain adaptation on both the ACDC dataset and the newly introduced ACG benchmark.

\section{Acknowledgement}
This work is supported by the Investitionsbank Berlin (IBB) (BerDiBa, grant no. 10185426) and by the German Federal Ministry for Economic Affairs and Climate Action (DeepTrain, grant no. 19S23005D).


%
%
%
%
\newpage
\bibliographystyle{splncs04}
\bibliography{017-main}

\end{document}